\documentclass[aps,prx,10pt,twocolumn,superscriptaddress,nofootinbib,floatfix]{revtex4-2}

\usepackage{graphicx}
\usepackage{dcolumn}
\usepackage{bm}
\usepackage{amsmath}

\usepackage{xcolor}
\usepackage{algorithm}%
\usepackage{algorithmicx}%
\usepackage{algpseudocode}%
\usepackage{braket}
\usepackage[normalem]{ulem}
\usepackage{amsmath,amssymb}
\usepackage{amsthm}

\usepackage[colorlinks=true,allcolors=blue]{hyperref}

\DeclareRobustCommand{\KOs}{\bgroup\markoverwith{\textcolor{teal}{\rule[0.5ex]{2pt}{0.9pt}}}\ULon}

\begin{document}

\preprint{APS/123-QED}

\title{Shot-budgeted Gibbs objective function-based training for QAOA as a sampler
}

\author{Aurélien Commans}
\author{Jérémie Messud}%
\affiliation{%
 TotalEnergies, 2 place Jean Millier, 92078 Paris La Défense Cedex, France}%

\date{\today}

\begin{abstract}

An increasingly studied feature of the Quantum Approximate Optimization Algorithm (QAOA) is its ability
to serve as a sampler.
Given a pragmatic shot budget $K$, the objective is to use QAOA to return with high probability at least one bitstring that improves upon a reference bitstring "to beat" obtained classically. From this perspective, the goal is not to train the QAOA output probability distribution until full concentration on the optimal bitstring(s), but to shape the distribution so that sufficient probability mass is
assigned to bitstrings whose energies lie below the energy of the reference bitstring. 
The standard (mean energy) QAOA objective function is not the best suited for such a task as it does not place a higher "weight" on the low-energy tail of the distribution. 
The conditional value at risk (CVaR) and Gibbs objective functions are recognized as better-suited but depend on hyper-parameters, $\alpha$ and $\eta$ respectively, whose values 
can strongly influence the efficiency of the training.
Fixing automatically these hyper-parameters from sound theoretical foundations represents an open question, and connecting these hyper-parameters to the given shot budget $K$ represents another one, both  being crucial for QAOA as a sampler.
In this article, we propose a method that resolves both problems, which is especially well suited to the Gibbs objective function (we also propose a first application to CVaR but the method is less adapted). It allows one to set $\eta$ automatically and robustly, at each optimizer call, without any prior knowledge of the problem. The method explicitly considers the given shot budget $K$ and allows one to control the accuracy of the objective function estimator (or equivalently the effective sample size). On weighted Max-Cut and portfolio-optimization
instances up to $25$ qubits and depths up to $p = 300$, and on simulator, we illustrate the efficiency of our ”shot-budgeted QAOA-Gibbs” scheme.

\end{abstract}

\maketitle

\section{Introduction}

The Quantum Approximate Optimization Algorithm (QAOA) is a variational
quantum algorithm originally proposed to solve combinatorial optimization
problems defined over $N$ binary variables, with each candidate solution encoded as a bitstring~\cite{farhi2014qaoa}. It has been later generalized within the Quantum Alternating Operator Ansatz framework~\cite{hadfield2019}, with the same acronym.
A depth-$p$ QAOA circuit alternates $p$ layers of a problem unitary (based on the problem Hamiltonian $H_P$) and a mixer unitary, resulting in $2p$ variational parameters to train. A
classical optimizer adjusts these parameters to minimize an objective
function, conventionally the mean "energy" denoted by $\left\langle H_P\right\rangle$, estimated from a number of bitstrings sampled from
the QAOA output probability distribution, these samples being also called QAOA measurements or shots.

Theoretical studies suggest that, at sufficiently large depth and provided suitable parameters are found, the QAOA output distribution becomes concentrated on the optimal bitstring(s) of the combinatorial problem. 
For certain problem families and suitable depth choice, theoretical and numerical studies further suggest that QAOA may achieve polynomial speedups over classical algorithms~\cite{boulebnane2024solving,shaydulin2024evidence,boulebnane2025evidence,apte2026quantum}, as well as stronger speedups (possibly exponential) in more restricted settings~\cite{golden2023numerical,farhi2025lower,montanaro2024quantum}.
Although these results do not establish a general quantum advantage for
combinatorial problems, they have made QAOA one of the  cornerstones and most studied variational quantum algorithms~\cite{zhou2020qaoa,basso2022highdepth,akshay2020reachability,larocca2022diagnosing,bittel2021nphard}.

In any case, the potential advantage of QAOA is conditional upon obtaining good 
parameters at an affordable cost, which remains one of
the potential bottlenecks of the algorithm. Indeed, parameter optimization is
itself a difficult problem.
QAOA (as any parameterized quantum circuit) can suffer
from barren plateaus, where gradients become exponentially small with the
system size $N$~\cite{mcclean2018barren}, as well as from rugged optimization
landscapes containing poor local minima
~\cite{bittel2021nphard,larocca2022diagnosing,rajakumar2024trainability}. Numerous training strategies have been
introduced to mitigate these difficulties, including layerwise training,
parameter initialization and transfer (leveraging parameter
concentration features), warm starts and interpolation between consecutive circuit depths
~\cite{zhou2020qaoa,egger2021warmstart}. One of the most advanced recent approach is the iterative interpolation scheme introduced in Ref.~\cite{apte2026iterative}. It allows one to control the smoothness of the QAOA schedules (i.e., the evolution of the QAOA parameters with depth) and their deviation from the adiabatic inspiration underlying QAOA by expanding them in a truncated polynomial basis. It also allows one to mitigate barren plateaus and local minima to some extent by progressively increasing both the circuit depth and the basis size
~\cite{apte2026iterative}.

Another important and increasingly studied feature of QAOA is its ability
to sample high-quality bitstrings, as measured by the objective function values (or "energies") of the combinatorial problem. Such a scheme is called QAOA as a sampler
~\cite{cepaite2025quantumenhancedoptimizationwarmstarts}.
Given a reasonable shot budget $K$, the goal is to use QAOA to return with high probability at least one bitstring that improves upon a reference bitstring we have at our disposal, here called bitstring "to beat". The latter can be obtained by a classical (heuristic) optimization of the combinatorial problem, the standard case in an industrial context, or a "worst acceptable" solution defined \textit{a priori}. From this perspective, the relevant goal is not to train the QAOA output probability distribution to convergence, i.e. until it fully concentrates on the optimal bitstring(s).
Rather, it is to shape its distribution so that sufficient probability mass is
assigned to bitstrings whose energies lie below the energy of the bitstring to beat, which is pragmatic from both training cost and operational relevance point of views. 
The mean energy objective function is not the best suited for such a task as it does not put a "higher weight" on the low-energy tail of the distribution. 
So, a way to foster the desired behavior is to train QAOA with more adapted objective functions.

Several works in this direction have studied how alternative objective functions can "favor" the low-energy tail of the QAOA output probability distribution~\cite{barkoutsos2020cvar,li2020gibbs,qiu2026}.
Among these, the conditional value at risk (CVaR) has been proposed in Ref.~\cite{barkoutsos2020cvar}.
It represents the energy mean over a
fraction $\alpha\in(0,1]$ of bitstrings having the lowest energies. Another option is the Gibbs objective function,  inspired by the free energy of statistical physics, 
and introduced for QAOA in Ref.~\cite{li2020gibbs}. It is defined as
    $-\frac{1}{\eta}
    \ln\left\langle e^{-\eta H_P}\right\rangle$ 
for $\eta>0$.
Whereas CVaR
discards the fraction $1-\alpha$ of the bitstrings with the highest energies,
the Gibbs objective function retains all bitstrings but exponentially
decreases the relative weight of high-energy ones depending on the value of $\eta$,
which leads to a better behavior in practice~\cite{li2020gibbs}.
Both CVaR and Gibbs objective functions depend on hyper-parameters, respectively $\alpha$ and $\eta$. Their values allow to control the "weight" placed on the low-energy tail
and thus can be crucial for the performance of QAOA as a sampler.
Previous works either fix these hyper-parameters using prior knowledge on the problem to solve \cite{li2020gibbs}, which is impractical, or make them follow a linear schedule~\cite{qiu2026,Kolotouros_2022} or set them heuristically \cite{barkoutsos2020cvar}.
Fixing automatically these hyper-parameters from sound theoretical foundations is highly desirable to ensure an efficient training of QAOA as a sampler and represents an open question. Connecting these hyper-parameters to a given QAOA shot budget $K$ represents another open question, as hyper-parameter values that produce a stronger weight on low-energy bitstrings make the QAOA objective function estimator dependent on a smaller effective number of shots, which can affect the training accuracy. 

In this article, we propose a method to resolve both problems, especially well suited to the Gibbs objective function. It allows one to set $\eta$ automatically and from clear theoretical considerations, at each optimizer call, without prior knowledge of the problem. The method explicitly considers a given shot budget $K$ and allows one to control the accuracy of the objective function estimator. Additionally, we propose a first application of our method to automatically set the CVaR $\alpha$ hyper-parameter, but demonstrate that CVaR is less suited to such a scheme. We then compare results by means of the QAOA output cumulative distribution function (CDF) of bitstrings sorted by non-decreasing energy, called energy-sorted CDF (eCDF), to evaluate the potential of our methods in a QAOA as a sampler context. We finally demonstrate numerically the interest of our "shot-budgeted QAOA-Gibbs".

The structure of the article is as follows:
\begin{enumerate}
    \item We introduce a straightforward QAOA as a sampler efficiency criterion based on the eCDF. This criterion is however not practical for QAOA training and is thus used as a benchmarking protocol.
    \item We justify further ahead the use of the Gibbs objective function to promote the desired
    behavior of the eCDF and propose a
    procedure to automatically set its
    hyper-parameter $\eta$ by means of a so-called $\chi^2$-budget rule. 
    At each optimizer call, the procedure solves
    $\widehat{\chi}^{\,2}(\eta)=B$ for $\eta$, where $B$ 
    is set to control an effective sample size.
    A similar procedure is set for the CVaR hyper-parameter $\alpha$, for the comparison. 
    \item We provide numerical evidence on the robustness of our QAOA with Gibbs scheme in comparison with standard QAOA and CVaR QAOA with $\alpha$ set by the $\chi^2$-budget rule, on weighted Max-Cut and portfolio
    optimization problems. The study is done on simulator for instances up to $25$ qubits and $p=32$. 
    \item We finally combine the $\chi^2$-budget rule with the iterative interpolation scheme~\cite{apte2026iterative}, introducing refinements designed to accelerate convergence, which allows us to do tests with much larger QAOA depth, up to $p=300$. 
    Interestingly, our method reaches a given low-energy concentration at smaller circuit depths, which could be particularly relevant in terms of resource optimization.
\end{enumerate}

\section{QAOA as a sampler and Gibbs objective function}

\subsection{Combinatorial optimization and QAOA}

Our goal is to solve the unconstrained combinatorial optimization problem
\begin{equation}
    \min_{z\in\{0,1\}^N} E_z,
    \label{eq:opt_pb}
\end{equation}
with an objective function $E_z$ that is a function of a bitsring $z$, here called an "energy" for consistency with QAOA wording.

The Max-Cut~\cite{goemans1995maxcut} or portfolio optimization~\cite{markowitz1952portfolio} problems, for instance, enter in such a category after some reformulations as we will see below. The search space over all possible bitstrings grows exponentially, with dimensionality $2^N$, so that an exhaustive search is not scalable classically. This leads to a worst-case complexity that is exponential for these problems, but does not prevent that many instances can be relatively easy to solve with classical optimizers up to large $N$ values.
General QAOA applications target hard instances, for which the best classical optimizers struggle to find a satisfying approximation of the global minimum even for reasonable $N$ values \cite{abbas2024quantum_optimization,koch2026qoblib}
(for practical reasons, our subsequent numerical results consider instances that act as a proxy for the hard regime).
More generally, we denote by $z_{i^\star}$ a bitstring "to beat", which can be obtained by classical optimization (standard in an industrial context) or can also consist in a "worst acceptable" solution defined \textit{a priori}.
We denote by $\{{z_0},{z_1},...,{z_{2^N-1}} \}$ the bitstrings sorted in increasing energy order, i.e. $E_{z_i}\leq E_{z_{i+1}}$.
In case of energy degeneracy, bitstrings are ranked by increasing binary to integer value. 
The promise of QAOA is to efficiently output bitstrings  $z$ such that $E_z \le E_{z_{i^\star}}$,
i.e. with resources that are polynomial in $N$.

The QAOA is a variational quantum algorithm  originally inspired from adiabatic quantum computing \cite{farhi2014qaoa,hadfield2019}. 
Its  ansatz is a $N$-qubits state given by 
\begin{align}
    \ket{\Psi(\boldsymbol{\gamma}, \boldsymbol{\beta})} = \prod_{k = 1}^pe^{-i\beta_kH_0}e^{-i\gamma_kH_P}\ket{\psi_0},
    \label{eq:ansatz}
\end{align}
where $\ket{\psi_0}$ is the QAOA initial state and $H_0$ is the "mixer" Hamiltonian.
Often, $H_0$ is chosen such that it can be easily implemented as a quantum circuit, and
$\ket{\psi_0}$ is chosen as the ground state of $H_0$, by analogy with adiabatic quantum computing.
The most common choice is $-H_0 = \sum_{i = 1}^{N}X_i$, where the Pauli operator $X_i$ acts on the $i-$th qubit,
and its ground state is the uniform superposition $\ket{\psi_0} = \frac{1}{\sqrt{2^N}}\sum_{z=0}^{2^N-1}\ket{z}$.
The "problem" Hamiltonian
\begin{align}
    H_P=\sum_{z=0}^{2^N-1} E_z\ket{z}\bra{z}
    \label{eq:HP}
\end{align}
is diagonal in the computational basis. It encodes the optimization problem of interest, eq. (\ref{eq:opt_pb}), and satisfies $E_{z}=\bra{z}H_P\ket{z}$.
For a given bitstring $z$, the QAOA output probability distribution is given by 
\begin{equation}
    P(z, \boldsymbol{\gamma}, \boldsymbol{\beta}) = \lvert \braket{z|\Psi(\boldsymbol{\gamma}, \boldsymbol{\beta})}\rvert^2.
\end{equation}

The depth $p$ of the QAOA circuit is a free parameter. 
There are $2p$ real parameters $(\boldsymbol{\gamma}, \boldsymbol{\beta})$, here called the schedules, that need to be optimized through a quantum-classical hybrid loop
such that the low energy bitstrings are related to high probabilities in $P(z, \boldsymbol{\gamma}^*, \boldsymbol{\beta}^*)$. 
The standard QAOA objective function for such a task considers the mean energy
\begin{equation}
    \min_{\boldsymbol{\gamma}, \boldsymbol{\beta}} \bra{\Psi(\boldsymbol{\gamma}, \boldsymbol{\beta})}H_P\ket{\Psi(\boldsymbol{\gamma}, \boldsymbol{\beta})}
    =\min_{\boldsymbol{\gamma}, \boldsymbol{\beta}}\sum_{z=0}^{2^N-1} P(z, \boldsymbol{\gamma}, \boldsymbol{\beta})E_z.
    \label{eq:std_QAOA}
\end{equation}

Another commonly used objective function to minimize is the conditional value at risk (CVaR)~\cite{barkoutsos2020cvar,rockafellar2000cvar}, which is the mean energy over 
the $\alpha \in (0, 1]$ probability mass on the lowest energies. 
It is defined by
\begin{align}
&\text{CVaR}_\alpha(\boldsymbol{\gamma}, \boldsymbol{\beta}) =
\frac{1}{\alpha}\Bigg[\sum_{z:\, E_z < E_{z_{i_\alpha(\boldsymbol{\gamma}, \boldsymbol{\beta})}}} P(z,\boldsymbol{\gamma}, \boldsymbol{\beta})\, E_{z} \nonumber \\
&\qquad + \Bigg(\alpha - \sum_{z:\, E_z < E_{z_{i_\alpha(\boldsymbol{\gamma}, \boldsymbol{\beta})}}} P(z,\boldsymbol{\gamma}, \boldsymbol{\beta})\Bigg) E_{z_{i_\alpha(\boldsymbol{\gamma}, \boldsymbol{\beta})}}\Bigg],
\label{eq:cvar}
\end{align}
where the second term ensures that exactly an $\alpha$ probability mass is included by correcting for any probability excess at the $\alpha$ threshold, defined by
\begin{align}
i_\alpha(\boldsymbol{\gamma},\boldsymbol{\beta}) = 
\min\left\{ i : \sum_{z:\,E_z \leq E_{z_i}} P(z,\boldsymbol{\gamma},\boldsymbol{\beta}) \geq \alpha \right\}.
\end{align}

Using the CVaR as a QAOA objective function can help to concentrate more probability mass on the low energy tail of the QAOA output distribution compared to the mean energy.
Indeed, contrariwise to standard QAOA potential situations,
it is not penalized for finding $(\boldsymbol{\gamma},\boldsymbol{\beta})$ schedules that put more probability mass on the low energy tail  (desired for QAOA as a sampler), below $E_{z_{i_\alpha(\boldsymbol{\gamma}, \boldsymbol{\beta})}}$, at the price of also putting more probability mass on higher energy bitstrings (undesired for the mean energy objective function).

Another interesting objective function for this purpose is the Gibbs one \cite{li2020gibbs,qiu2026}
\begin{align}
F_\eta &= -\frac{1}{\eta}\ln \bra{\Psi(\boldsymbol{\gamma}, \boldsymbol{\beta})} e^{-\eta H_P}\ket{\Psi(\boldsymbol{\gamma}, \boldsymbol{\beta})} \nonumber\\
&= -\frac{1}{\eta}\ln \sum_{z=0}^{2^N-1} P(z,\boldsymbol{\gamma}, \boldsymbol{\beta})\,e^{-\eta E_z},
\label{eq:gibbs}
\end{align}
with $\eta >0$ the so-called tilt parameter. Minimization of this objective function for sufficiently large $\eta$ values also rewards concentration of probabilities on low energy bitstrings. It has no hard cutoff contrariwise to CVaR.

From now on we drop the $(\boldsymbol{\gamma}, \boldsymbol{\beta})$ from the notations for simplicity, which means next reasoning are conditioned to given $(\boldsymbol{\gamma}, \boldsymbol{\beta})$ values.

\subsection{Energy-sorted CDF and shot budget for QAOA as a sampler}
\label{sec:qaoa_theory}

We are interested in the energy-sorted CDF (eCDF), which we define as the CDF of the QAOA output probability distribution sorted by energy 
\begin{align}
    \text{eCDF}(E) = \sum_{z: E_z\leq E} P(z) = P(\{z: E_z\leq E\}),
    \label{eq:eCDF}
\end{align}
that represents a meaningful measure of the QAOA output probability distribution  "concentration".
Specifically, the value $\text{eCDF}(E_{z_{i^\star}})$ represents the probability that a single QAOA shot provides a bitstring of better quality than (or equivalent to) the bitstring to beat $z_{i^\star}$.

\begin{figure}[ht]
    \centering   
    \includegraphics[width=0.9\linewidth]{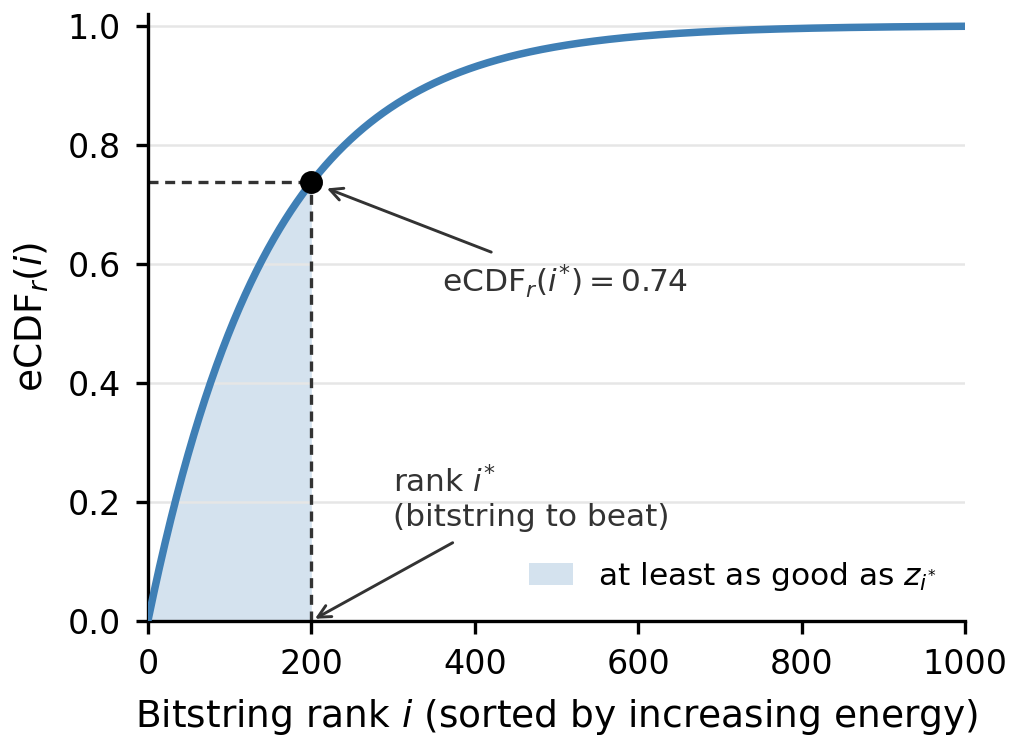}
    \caption{eCDF over the bitstrings sorted in increasing energy. More precisely, the "rank" eCDF defined by $\text{eCDF}_r(i) = \sum_{j=0}^{i}P(z_j)$ is shown.}
    \label{fig:ecdf}
\end{figure}

Now, let us dive into the QAOA as a sampler case.
Given a reasonable shot budget $K$, that is low-order polynomial in $N$ and represents the maximum number of QAOA shots we can afford, the goal is to sample high-quality bitstrings $z$ as measured by $E_z$.
More precisely, the goal is to train the QAOA output probability distribution such that sufficient probability mass is assigned to bitstrings whose energies lie below the energy of the bitstring to beat, i.e. such that $\text{eCDF}(E_{z_{i^\star}})$ is high~\cite{cepaite2025quantumenhancedoptimizationwarmstarts}.
An illustrative example is given in Fig.~\ref{fig:ecdf}. 

Assuming that the $K$ shots are independent and identically distributed according to the QAOA output probability distribution, 
the probability of obtaining at least one bitstring that beats $z_{i^\star}$ after $K$ shots is $P_{\mathrm{success}}=1-\left(1-\text{eCDF}(E_{z_{i^\star}})\right)^K$. Having
\begin{equation}
    P_{\mathrm{success}}\ge 1-\delta,
\end{equation}
implies to reach an eCDF concentration that satisfies
\begin{equation}
    \text{eCDF}(E_{z_{i^\star}}) \geq 1-\delta^{1/K}.
    \label{eq:concentration}
\end{equation}
This gives a potential criterion for the  training of QAOA as a sampler:
Given a bitstring to beat and the corresponding $E_{z_{i^\star}}$, a maximal shot budget $K$ we can afford
and a reasonable probability of success $1-\delta$,
we might consider the training successful
when the concentration criterion in eq. (\ref{eq:concentration}) is reached.
Then, the probability to measure at least one bitstring whose energy is less or equal than the energy of the bitstring to beat with $K$ QAOA shots is at least $1-\delta$.

A partially open question is if eq.~(\ref{eq:concentration}) can be satisfied for a given $1-\delta$ value and a shot budget $K\ll 2^N$ that is low-order polynomial in $N$, while keeping the circuit depth $p$ and training cost tractable, despite the exponential growth of the $2^N$-cardinality set indexing $P(z)$.
This question goes beyond the scope of this work but we mention that certain properties of QAOA provide motivation for investigating this  direction~\cite{farhi2014qaoa,lotshaw2023boltzmann,DiezValle2026}~
\footnote{
For unweighted MaxCut on regular graphs of fixed degree, QAOA at fixed $p$ has a measured-energy variance bounded by $O(N)$, implying concentration of the energy around its mean~\cite{farhi2014qaoa}. Also, for globally optimized standard QAOA, increasing $p$ guarantees that the mean energy does not increase and tends to $E_{z_0}$ as $p\to\infty$, so that the lowest energy bitstrings  probability 
tends to one~\cite{farhi2014qaoa}.
}.

As the eCDF can never be completely known in practice, we must work with its estimator, which is given for $K$ shots by 
\begin{align}
    &\widehat{\text{eCDF}}(E_{z_{i^\star}})= \frac{1}{K}\sum_{k=1}^K \mathbf{1}_{(E(k) \le E_{z_{i^\star}})},
    \label{eq:eCDFestimator}
\end{align}
where $E(k)$ represents the energy related to the bitstring $z(k)$ measured by shot $k$ ($E(k)=E_{z(k)}$).
In general, the number of shots used for the estimators could be different from the number of shots $K$ used to sample high quality bitstrings once the QAOA trained.
However and even if there is no obligation to do so, as $K$ represents our maximum shot budget at fixed $(\boldsymbol{\gamma}, \boldsymbol{\beta})$, we here keep the same number for the estimators.

Let us imagine a QAOA training based on
\begin{equation}
    \max_{\boldsymbol{\gamma}, \boldsymbol{\beta}} \, \widehat{\text{eCDF}}(E_{z_{i^\star}}),
    \label{eq:eCDF_ill}
\end{equation}
stopped when $\widehat{\text{eCDF}}(E_{z_{i^\star}})$ does not improve anymore or gets larger than $1-\delta^{1/K}$,
which should theoretically lead to better concentration properties than these with the standard QAOA scheme
\footnote{Indeed, the mean energy objective is penalized for putting probability mass on high energy bitstrings, which can potentially hamper concentration of probabilities on the low energy tail of the probability distribution. For example, for 2 qubits sorted by non decreasing energy $\{z_0,z_1,z_2, z_3\}$ and $i^\star = 1$, the distribution $\{0.3,0.3,0,0.4\}$ is better than $\{0.3,0.2,0.5,0\}$ in the eCDF framework, while possibly having a higher mean energy (if $E_{z_3} - E_{z_2}$ is large).}.
However, such an eCDF-based scheme
would be ill-posed. 
Firstly, if the amount of bitstrings of energy below $E_{z_{i\star}}$ is relatively small and $K$ is polynomial in $N$ (leading to $i^\star+1 \ll 2^N/K$), as it is the case in practical applications, it becomes impossible to estimate the eCDF properly at the start of the QAOA training that considers a quasi-uniform distribution $P(z)\approx 1 / 2^N$. 
Secondly, different $P(z_j)$ values for $E_{z_j}\le E_{z_{i^\star}}$ can lead to the same $\text{eCDF}(E_{z_{i^\star}})$ value at fixed $E_{z_{i^\star}}$.

As a consequence, we will consider in the following an eCDF maximization compliant objective function that removes the ill-posedness and the direct dependence on $E_{z_{i^\star}}$.
The eCDF could still be used as a benchmark to control a posteriori the quality of the solutions 
obtained through our scheme, or used as an additional measure monitored during the QAOA training (one might still decide to stop whether $\widehat{\text{eCDF}}(E_{z_{i^\star}})$ does not improve anymore or just larger than $1-\delta^{1/K}$).

\subsection{Gibbs as a practical objective function and remaining problem}

We start from the following identity, demonstrated in Appendix~\ref{app:proof-ecdf},
\begin{align}
&\forall\eta>0:\quad \sum_{z=0}^{2^N-1} P(z)\,e^{-\eta (E_z-E_{z_0})} \nonumber\\
 &\quad= \int_0^\infty \eta e^{-\eta x}\,\mathrm{eCDF}(E_{z_0}+x)\,dx \nonumber\\
 &\quad= \mathbb{E}_{X\sim\mathrm{Exp}(\eta)}\!\left[\mathrm{eCDF}(E_{z_0}+X)\right],
\label{eq:lemma_ecdf}
\end{align}
where $\mathrm{Exp}(\eta)$ is defined through its  probability density $\rho(x)=\eta e^{-\eta x}\mathbf{1}_{(x\geq 0)}$ and has mean $1/\eta$.
The left hand-side of eq.~(\ref{eq:lemma_ecdf}) can thus also be interpreted as a weighted mean of the eCDF over all energy thresholds $E_{z_0}+x$ for $x\ge 0$, with a weight that favors the low-energies. 
The hyper-parameter $\eta>0$ controls the strength of the weighting towards the low energies.

Considering an objective function based on the maximization of eq.~(\ref{eq:lemma_ecdf}) is thus equivalent to maximize the eCDF mean over a random threshold $E_{z_0}+X$,
where $X$ is drawn from an exponential distribution with mean $1/\eta$. 
It can  be considered as being related to the maximization of the eCDF concentration over the low energies while removing
the ill-posedness that would be related to the direct maximization of the eCDF at a given value $E_{z_{i^\star}}$ (recall the considerations in the previous section).
Indeed, as there is no hard threshold $E_{z_{i^\star}}$ in eq.~(\ref{eq:lemma_ecdf}), all shots  should contribute to its estimator.
Secondly, as all eCDF values are considered in the expectation  of eq.~(\ref{eq:lemma_ecdf}) (and not only the eCDF value at fixed $E_{z_{i^\star}}$), 
moving probability towards a lower-energy bitstring always increases the score, contrariwise to the eCDF at fixed $E_{z_{i^\star}}$ that can be insensitive to such a move in some situations.

Still, eq.~(\ref{eq:lemma_ecdf}) depends on $E_{z_0}$, the solution of the combinatorial problem.
However, this dependency reduces to a constant factor $e^{\eta E_{z_0}}$ that may be omitted as from the perspective of a usage of this equation as an objective function.
As a consequence, we might consider to train QAOA by maximization of
\begin{align}
\label{eq:gibbs_sum}
\sum_{z=0}^{2^N-1} P(z)\,e^{-\eta E_z} &= \bra{\Psi} e^{-\eta H_P}\ket{\Psi},
\end{align}
keeping the eCDF-associated interpretation related to eq.~(\ref{eq:lemma_ecdf}).

Note that the eq.~(3) from Ref. \cite{li2020gibbs} translates in our notations and a fixed threshold $E_{z_{i^\star}}$ into
\begin{align}
\forall\eta>0: 
\mathrm{eCDF}\!\left(E_{z_{i^\star}}\right) &= \sum_{z=0}^{2^N-1} P(z)\, \mathbf{1}_{(E_z\leq E_{z_{i^\star}})}
\nonumber\\
&\leq \sum_{z=0}^{2^N-1} P(z)\, e^{-\eta(E_z-E_{z_{i^\star}})},
\end{align}
from which we deduce, using eq.~(\ref{eq:lemma_ecdf})
\begin{align}
\label{eq:ecdf-gibbs}
\mathrm{eCDF}\!\left(E_{z_{i^\star}}\right) \leq
e^{\eta E_{z_{i^\star}}}\bra{\Psi} e^{-\eta H_P}\ket{\Psi}.
\end{align}
Eq.~(\ref{eq:ecdf-gibbs}) provides an upper bound on $\mathrm{eCDF}(E_{z_{i^\star}})$ and thus a necessary condition for reaching a prescribed $\mathrm{eCDF}(E_{z_{i^\star}})$ value, but it does not by itself justify maximizing the objective in eq.~(\ref{eq:gibbs_sum}).
A complementary lower bound can be derived by separating the contributions $E_z\leq E_{z_{i^\star}}$ and $E_z>E_{z_{i^\star}}$  in eq.~(\ref{eq:gibbs_sum}), which yields 
$\langle\Psi|e^{-\eta H_P}|\Psi\rangle \leq e^{-\eta E_{z_0}} \mathrm{eCDF}(E_{z_{i^\star}}) + e^{-\eta E_{z_{i^\star}}} [1-\mathrm{eCDF}(E_{z_{i^\star}})]$. We obtain, for $E_{z_{i^\star}}>E_{z_0}$
\begin{equation} 
\frac{ e^{\eta E_{z_{i^\star}}}\bra{\Psi} e^{-\eta H_P}\ket{\Psi} - 1}{ e^{\eta (E_{z_{i^\star}} -E_{z_0})}-1 }
\leq
\mathrm{eCDF}(E_{z_{i^\star}}).
\label{eq:ecdf-gibbs2}
\end{equation} 
Eq. (\ref{eq:ecdf-gibbs2}) provides a lower bound on $\mathrm{eCDF}(E_{z_{i^\star}})$ which is a strictly increasing function of the objective in eq.~(\ref{eq:gibbs_sum}) and thus justifies maximizing eq.~(\ref{eq:gibbs_sum}).

Finally, we note that the maximization of eq. (\ref{eq:gibbs_sum}) is equivalent to the minimization of the Gibbs objective function~\cite{li2020gibbs,qiu2026}, eq.~(\ref{eq:gibbs}), which we will consider in the following. 
An interest of the Gibbs objective function compared to the objective function in eq. (\ref{eq:gibbs_sum}) is that it corresponds asymptotically to two different regimes~\cite{qiu2026,li2020gibbs}
\begin{align}
&\lim_{\eta \rightarrow 0^+}F_\eta = \bra{\Psi}H_P\ket{\Psi}, \nonumber\\
&\lim_{\eta \rightarrow\infty}F_\eta =E_{z_0},
\label{eq:lim_eta}
\end{align}
the last relationship being valid if $P_0 >0$, where 
\begin{align}
P_0=\sum_{z:E_z=E_{z_0}}P(z),
\label{eq:P0}
\end{align}
is the probability of the lowest energy level.
Eq. (\ref{eq:lim_eta}) means that $\eta$ plays for the Gibbs objective function a similar role than the one $\alpha$ plays for CVaR in eq.~(\ref{eq:cvar}). Indeed, $E_{z_0}\le F_\eta\le\bra{\Psi}H_P\ket{\Psi}$ for all $\eta$ by Jensen's inequality and $F_\eta$ is non-increasing in $\eta$, just as CVaR decreases from $\bra{\Psi}H_P\ket{\Psi}$ at $\alpha=1$ to $E_{z_0}$ for $\alpha\le P_0$. 
However, in the $\alpha \rightarrow 0$ limit, CVaR no longer depends on  $(\boldsymbol{\gamma}, \boldsymbol{\beta})$ and is not rewarded for increasing $P_0$. But for $\eta$ much larger than the inverse of the first energy gap, we have $F_\eta\simeq E_{z_0}-\ln(P_0)/\eta$ and minimizing $F_\eta$ amounts to maximizing $P_0$.
For intermediate $\eta$, the Gibbs objective function is quite different from the standard QAOA or CVaR objective functions,
fostering a better eCDF concentration towards the low energies in practice.
Finally, unlike CVaR (or the eCDF), Gibbs has no hard cutoff; it gives every bitstring a
weight that decreases smoothly with the corresponding energy. 

The considerations above justify the minimization of the Gibbs objective function as a well-posed optimization problem that tends to enhance the eCDF-based concentration towards the low energies, thus suited to the training of QAOA as a sampler.
This has been here justified from a complementary point of view than in Ref. \cite{li2020gibbs} and is valid if pertinent $\eta$ values can be defined, that we will discuss later.

In practice and for a shot budget $K$, the Gibbs objective function estimator must be used
\begin{align}
    \label{eq:Gibbs_estimator}
    \hat F_\eta = -\frac{1}{\eta} \ln \frac{1}{K}\sum_{k = 1}^K e^{-\eta E(k)}.
\end{align}
In comparison, the standard QAOA objective function estimator is given by
\begin{align}
    \label{eq:mean_estimator}
    \widehat{\bra{\Psi}H_P\ket{\Psi}} = \frac{1}{K}\sum_{k=1}^{K} E(k),
\end{align}
and the CVaR QAOA objective function estimator by
\begin{align}
\label{eq:CVAR_estim}
\widehat{\text{CVaR}}_\alpha = \frac{1}{\lceil K\alpha \rceil}\sum_{i=1}^{\lceil K\alpha \rceil} E_{[i]},
\end{align}
where $E_{[1]}\le\dots\le E_{[K]}$ are the $K$ measured energies sorted in increasing order, and $E_{[\lceil \alpha K \rceil]}$ is the estimator of the value at risk $\mathrm{VaR}_\alpha$~\cite{rockafellar2000cvar,rockafellar2002cvar}.

{The remaining problem consists in setting $\eta$ automatically, which should be somewhat related to the given shot budget $K$ through the Gibbs objective function estimator, eq. (\ref{eq:Gibbs_estimator}).}
Prior work fixes $\eta$ a priori \cite{li2020gibbs} or makes it follow a linear schedule~\cite{qiu2026}, but setting it automatically without prior knowledge of a specific instance remains an open question as mentioned in Ref.~\cite{qiu2026}, important for the success of a Gibbs based QAOA as a sampler scheme.

\section{\texorpdfstring{$\chi^2$}{chi-square}-divergence based rule and automated computation of the Gibbs objective function}
\label{sec:chi2}

\subsection{Variance and \texorpdfstring{$\chi^2$}{chi-square}-divergence}

We would like to quantify the variance of the estimators of the objective functions used within QAOA, eqs.~(\ref{eq:Gibbs_estimator})-(\ref{eq:CVAR_estim}), 
when various experiments with $K$ shots are done.
The basis of the method we will develop is general but the application will be particularly interesting when applied to the Gibbs objective function estimator, eq.~(\ref{eq:Gibbs_estimator}).

The intuition behind our proposal is that the variance of the estimators of a weighted objective function evaluated from $K$ samples of the QAOA output distribution $P(z)$ gives a measure of the quality of the estimator, which we ideally would like to control during training.
Another measure of quality could be the "effective sample size" \cite{kish1965survey},
 related to the fact that a weighted objective function lets some shots count more than others, leading to fewer effective shots and a less accurate objective function estimator.

In the case of the Gibbs objective function, we will show that the two are directly related: To the leading order and at a given number of shots $K$, the variance of the estimator of $\eta F_\eta$ is fixed by the effective sample size. It is then possible to keep the quality of the objective function estimator approximately constant, which gives a robust and automatic way to set the hyper-parameter $\eta$ during the training.

The rule we will develop is close to adaptive tempering in sequential Monte Carlo, where the next inverse temperature is chosen so that the effective sample size of the incremental weights is a fixed fraction of the number of samples~\cite{jasra2011inference,beskos2016adaptive}.

It is convenient to describe objective functions mentioned above using a reweighted distribution and its
normalization factor,
\begin{align}
Q_w(z) = \frac{P(z)\,w_z}{Z_w}, \quad Z_w = \sum_{z=0}^{2^N-1} P(z)\,w_z=\mathbb{E}_P\left[w\right],
    \label{eq:reweight}
\end{align}
for non-negative weights $w_z$, that must also preserve the invariance of the objective functions by translation ($H_P\rightarrow H_P + c$).
In the following and as we will make it clear, $Z_w$ will represent an objective function. Its estimator defined by
\begin{align}
    \hat Z_w = \frac{1}{K}\sum_{k=1}^K w(k),
\end{align}
where $w(k)$ represents the weight related to the bitstring $z(k)$ measured by shot $k$ (energy $E(k)$).

For $K$ independent and identically distributed shots drawn from $P(z)$, the estimator $\hat Z_w$ has variance
\begin{align}
\mathrm{Var}(\hat Z_w) = \frac{\mathrm{Var}_P(w)}{K} 
&= \frac{\mathbb{E}_P\left[w^2\right] - \mathbb{E}_P\left[w\right]^2}{K} \nonumber\\
&= \frac{\mathbb{E}_P\left[w^2\right] - Z_w^2}{K}.
\end{align}
As a consequence, when $Z_w\ne 0$ the relative variance of $\hat Z_w$ is 
\begin{equation}
    \frac{\mathrm{Var}(\hat Z_w)}{Z_w^2} = \frac{\chi^2(Q_w \Vert P)}{K},
\label{eq:relvar}
\end{equation}
where the $\chi^2$-divergence  \cite{sason2016fdivergence,agapiou2017importance} has been introduced
\begin{equation}
    \chi^2(Q_w \Vert P)  = \frac{\mathbb{E}_P\left[w^2\right]}{Z_w^2} - 1=\frac{\operatorname{Var}_P(w)}{Z_w^2}.
\label{eq:chi2}
\end{equation}
The $\chi^2$-divergence measures how far the reweighted distribution $Q_w$ in eq. (\ref{eq:reweight}) used in the QAOA objective function is from the QAOA output distribution $P$
\footnote{The more $Q_w$ differs from $P$, the larger $\chi^2(Q_w \Vert P)$ is. $\chi^2(Q_w \Vert P)$ vanishes only when
$Q_w=P$, which happens only if $w_z = Z_w$ is a constant weight over all bitstrings with $P(z) \ne 0$.}.
{The variance and relative variance of $\hat Z_w$ are related to the quality of the estimator of the weighted objective function evaluated from samples of $P$. The result above demonstrates that this variance gets higher as
the discrepancy between  the reweighted distribution $Q_w$ and $P$ as measured by the $\chi^2$-divergence grows, i.e. as the weights become less uniform. 
In other terms, the $\chi^2$-divergence quantifies the statistical loss induced by a non-uniform reweighting, which is related to the concept of effective sample size \cite{kong1994sequential,agapiou2017importance}.

Using the fact that the variance is invariant under translation (by $E_{z_0}$ or $\mathrm{VaR}_\alpha$ according to the case), we now set:
\begin{itemize}
    \item For standard QAOA:
\begin{align}
\mathrm{Var}\left(\widehat{\bra{\Psi}H_P\ket{\Psi}}\right)&=\mathrm{Var}(\hat Z_{w})=Z_{w}^2\frac{\chi^2(Q_{w} \Vert P)}{K} \nonumber\\
&=\frac{\operatorname{Var}_P(w)}{K}, \nonumber\\
&\hspace{-2.6cm} w_z=E_z-E_{z_0}.
\label{eq:var_mean}
\end{align}
\item For CVaR QAOA:
\begin{align}
\mathrm{Var}(\widehat{\operatorname{CVaR}}_\alpha)&\simeq \frac{1}{\alpha^2}\mathrm{Var}(\hat Z_{w})=\frac{1}{\alpha^2}Z_{w}^2\frac{\chi^2(Q_{w} \Vert P)}{K} \nonumber\\
&=\frac{1}{\alpha^2}\frac{\operatorname{Var}_P(w)}{K}, \nonumber\\
&\hspace{-1.8cm}  w_z= (\mathrm{VaR}_\alpha-E_z) \times  \mathbf{1}_{(E_z\leq \mathrm{VaR}_\alpha)}.
\label{eq:var_cvar}
\end{align}
    \item For Gibbs QAOA:
\begin{equation}
\begin{split}
\mathrm{Var}(\hat F_\eta)&\simeq\frac{1}{\eta^2}\frac{\mathrm{Var}(\hat Z_w)}{Z_w^2}=\frac{1}{\eta^2}\frac{\chi^2(Q_\eta\Vert P)}{K}, \\
&\hspace{-1.1cm} w_z=e^{-\eta (E_z-E_{z_0})},
\end{split}
\label{eq:var_Gibbs}
\end{equation}
where the variance in the last equation is approximated to leading order at a fixed $\eta$ by the delta method~\cite{oehlert1992delta}, using $\delta\ln\hat Z_\eta\simeq\delta\hat Z_\eta/Z_\eta$ as $\hat F_\eta=-(\ln\hat Z_\eta)/\eta$ depends on the shots only through $\hat Z_\eta$~
\footnote{
We have $\lim_{\eta\rightarrow 0^+} \mathrm{Var}(\hat F_\eta)=\mathrm{Var}(\widehat{\bra{\Psi}H_P\ket{\Psi}})$, which is consistent with the first line of eq. (\ref{eq:lim_eta}).
Indeed, for the Gibbs weights $\chi^2(Q_w\Vert P) = \eta^2\operatorname{Var}_P(E)+O(\eta^3)$,
where 
$\operatorname{Var}_P(E) = \sum_{z=0}^{2^N-1} P(z)E_z^2 - (\sum_{z=0}^{2^N-1} P(z)E_z)^2 = \sum_{z=0}^{2^N-1} P(z)(E_z-E_{z_0})^2 - (\sum_{z=0}^{2^N-1} P(z)(E_z-E_{z_0}))^2$ is invariant by translation. 
The variance of the Gibbs objective function 
is thus
$
    \mathrm{Var}(\hat F_\eta)\simeq\frac{1}{\eta^2}\frac{\chi^2(Q_\eta\Vert P)}{K}=\frac{1}{K}\operatorname{Var}_P(E)+O(\eta)
    \label{eq:lim_var_Gibbs}
$,
and the variance of the mean energy objective function estimator can be written
$
\mathrm{Var}(\widehat{\bra{\Psi}H_P\ket{\Psi}})=\frac{1}{K}\operatorname{Var}_P(E)\nonumber
$,
leading to the desired relation.
}.

\end{itemize}

Interestingly, only the Gibbs objective function variance depends directly on $\chi^2(Q_w\Vert P)$, which represents the object on which we will work now.

\subsection{\texorpdfstring{$\chi^2$}{chi-square}-divergence and effective sample size}

The effective sample size (defined as in Ref. \cite{owen2013mc} for un-normalized weights and $K$ shots)
\begin{equation}
    \label{ea:chiestimator_Keff}
    K_{\mathrm{eff}}(w) = \frac{K}{1+\chi^2(Q_w \Vert P)},
\end{equation}
quantifies the number of "weighted" shots drawn from $Q_w$ 
that carry an equivalent amount of statistical information than "uniformly weighted" shots drawn for $P$. It is an indicator related to the statistical loss induced by the non-uniform reweighting $w$. The larger the mismatch between $Q_w$ and $P$ measured by the $\chi^2$-divergence, the smaller $K_{\mathrm{eff}}(w)$ is.

In practice, we use the estimator of the $\chi^2$-divergence
\begin{equation}
    \hat\chi^2(Q_w \Vert P) = K\,\frac{\sum_{k=1}^K w(k)^2}{\left[\sum_{k=1}^K w(k)\right]^2} - 1,
    \label{ea:chiestimator}
\end{equation}
leading to the effective sample size estimator
\begin{equation}
    \hat K_{\mathrm{eff}}(w) = \frac{\left[\sum_{k=1}^K w(k)\right]^2}{\sum_{k=1}^K w(k)^2},
    \label{ea:chiestimator2}
\end{equation}
which can be obtained by computing $w(k)$ from each shot. 

The effective sample size estimator is not perfect. As $\hat K_{\mathrm{eff}}(w)$ and $\hat\chi^2(Q_w \Vert P)$ are estimated from $K$ shots, we keep $B \ll K$ so that $K_{\mathrm{eff}} = K/(1+B)$ remains large.
Note that it has been shown in Ref.~\cite{kish1965survey}  that another  more commonly used definition of the effective sample size, called} the "effective sample size of a reweighting",
rests on several approximations, does not depend on the quantity being estimated, and can be misleading, notably for rare events. Our $\hat K_{\mathrm{eff}}$ is built for the quantity being estimated, $Z_w$, as suggested in Ref.~\cite[Sec.~5.1]{elvira}, but is still computed from the same $K$ shots and bounded by $K$
\footnote{
Indeed, the $\chi^2$-divergence estimator is bounded, $0 \leq \hat\chi^2(Q_w \Vert P) \leq K-1$. The lower bound holds because a variance is non-negative, and the upper bound holds because the weights are non-negative.
}.
For the Gibbs objective function specifically, the weights happen to be the same in both definitions.

\subsection{$\chi^2$-budget rule, automated computation of \texorpdfstring{$\eta$}{eta} and application to CVaR}
\label{sec:rule_Gibbs_CVaR}

As explained above, the estimator $\hat\chi^2(Q_w \Vert P)$ can be computed from the measured bitstrings (or shots)
and used to define an effective sample size estimator $\hat K_{\mathrm{eff}}(w)$ through eq. (\ref{ea:chiestimator_Keff}).
For practical purpose, we could want to keep $\hat K_{\mathrm{eff}}(w)$ constant
\begin{equation}
    \hat K_{\mathrm{eff}}(w) = \frac{K}{1+B}, \qquad B > 0,
\end{equation}
and find the objective function weights $w$ within a pre-defined family of weights that satisfy
\begin{equation}
    \hat\chi^2(Q_w \Vert P) = B.
    \label{eq:rule}
\end{equation}
We then obtain weights $w(B)$ for which $K$ measured shots always carry a constant $\hat K_{\mathrm{eff}}(w) = K/(1+B)$ effective number of shots
from the point of view of the mismatch between $Q_w$ and $P$ as measured by the $\chi^2$-divergence.

{This represents an interesting potential change of point of view in case of objective functions parameterized by a hyper-parameter, like Gibbs. The hyper-parameter defines a family of weights.}
We then could fix how many effective shots one would like (through the $K$ and $B$ values) and adapt accordingly the hyper-parameter (that parameterize the weights) by solving eq. (\ref{eq:rule}).
So, rather than fixing a parameter of the objective, or ramp it a priori, an adaptive scheduling can be computed automatically at each optimizer call.

For the particular case of the Gibbs objective function, eq. (\ref{eq:var_Gibbs}) shows that controlling the effective sample size $K_{\mathrm{eff}}(w)$ is directly related to controlling the variance of the Gibbs objective function estimator (remind that the dependency in $E_{z_0}$ cancels out)
\begin{align}
\mathrm{Var}(\hat F_\eta)&\simeq\frac{\chi^2(Q_\eta\Vert P)}{K\eta^2}\simeq\frac{\hat\chi^2(Q_\eta\Vert P)}{K\eta^2} \nonumber\\
&=\frac{1}{\eta^2}\left(\frac{1}{\hat K_{\mathrm{eff}}} - \frac{1}{K}\right)=\frac{B}{K\eta^2}, \nonumber\\
&\quad w_\eta(z)=e^{-\eta E_z},
\label{eq:var_Gibbs2}
\end{align}
where
\begin{equation}
    \hat K_{\mathrm{eff}}(\eta) = \frac{\left[\sum_{k=1}^K e^{-\eta E(k)}\right]^2}{\sum_{k=1}^K e^{-2\eta E(k)}},
    \label{eq:Gibbs_solve}
\end{equation}
with $Z_\eta = \sum_{z=0}^{2^N-1} P(z)\, e^{-\eta E_z}$ and $ Q_\eta(z) =P(z)\, e^{-\eta E_z}/Z_\eta$.
As the weights are parameterized by $\eta$, solving eq. (\ref{eq:rule}) for a fixed $B$,
\begin{equation}
    \hat\chi^2(Q_\eta\Vert P)=K\,\frac{\sum_{k=1}^K e^{-2\eta E(k)}}{\left[\sum_{k=1}^K e^{-\eta E(k)}\right]^2}-1=B,
    \label{eq:rule_eta}
\end{equation}
or equivalently $\hat K_{\mathrm{eff}}(\eta)=K/(1+B)$, allows in principle to set the hyper-parameter $\eta$ automatically at each optimizer call.
Finally, note that the $\chi^2$-divergence increases monotonically (and continuously) with $\eta$ in the Gibbs case.

The optimizer is given $\hat F_\eta$ a an input (Alg.~\ref{alg:constant-depth}), with $\eta$ kept fixed during a call. At the angles where $\eta$ is set, and when eq.~(\ref{eq:rule}) is satisfied, $\eta\hat F_\eta=-\ln\hat Z_\eta$ has a standard deviation of approximately $\sqrt{B/K}$ at every call (i.e.\ $\sqrt{B/K}/\eta$ for $\hat F_\eta$) and a bias, positive by Jensen's inequality, of about $B/(2K)$.
Due to the first-order approximation at a fixed $\eta$ on the logarithm inherent to eq. (\ref{eq:var_Gibbs}) and the $\chi^2$-divergence or effective sample size estimators limitations due to rare outcomes absent from the samples of $P$, the $\eta\hat F_\eta$ estimate should be understood as an approximation requiring both sufficiently accurate estimation of the relevant weight moments and sufficiently small relative fluctuations of $\widehat{Z}_{\eta}$. Beyond these situations, the proposed rule becomes an empirical calibration criterion.  

Interestingly, the $\chi^2$-divergence also appears in eqs.~(\ref{eq:var_mean})-(\ref{eq:var_cvar}), i.e. in the context of the mean energy and CVaR objective functions estimators. However, we see that the $\chi^2(Q_w \Vert P)$ (or $K_{\mathrm{eff}}(w)$) value is not anymore directly related to the variance of the objective function estimator as a multiplication by $Z_w$ occurs.
In the case of the mean energy objective, there is no hyper-parameter of the objective that allows one to control the variance additionally to $K$ \cite{Kim2025DDS}.

In the case of CVaR, our formalism might also be used to set the hyper-parameter $\alpha$ automatically. Even if this is not the main objective of the paper, we mention for completeness that two possibilities occur:
\begin{itemize}
        \item Impose the same $\chi^2$ budget $B$ as for Gibbs at each optimizer call, which means controlling the effective sample size as an accuracy condition. With $\widehat{\mathrm{VaR}}_\alpha$ and $\widehat{\operatorname{CVaR}}_\alpha$ estimated from the $K$ shots, the weights of eq. (\ref{eq:var_cvar}) lead to
    \begin{align}
    1+\hat\chi^2(Q_w\Vert P)=\frac{1}{\alpha}\frac{\hat\sigma_\alpha^2}{(\widehat{\mathrm{VaR}}_\alpha- \widehat{\operatorname{CVaR}}_\alpha)^2},
\label{eq:chi2_cvar}
    \end{align}
    where $\hat\sigma_\alpha^2=\frac{1}{K\alpha}\sum_{k:  E(k)\le \widehat{\mathrm{VaR}}_\alpha} (\widehat{\mathrm{VaR}}_\alpha - E(k))^2$ is the mean squared distance to the value at risk of the energies below it, so that
    \begin{equation}
    \label{eq:Keff_cvar}
    \hat K_{\mathrm{eff}}(\alpha)=K\alpha\frac{(\widehat{\mathrm{VaR}}_\alpha- \widehat{\operatorname{CVaR}}_\alpha)^2}{\hat\sigma_\alpha^2}.
    \end{equation}
    Unlike the Gibbs case, the $\chi^2$-divergence is not continuous in $\alpha$, it cannot increase as $\alpha$ grows,
    and it changes only when $\widehat{\mathrm{VaR}}_\alpha$ crosses an energy level. Eq. (\ref{eq:rule}) then has no solution in general, and a $\chi^2$-divergence budget gives $\widehat{\mathrm{VaR}}_\alpha$ rather than $\alpha$ itself. We therefore take the smallest $\alpha$ for which $\hat\chi^2(Q_w\Vert P)\le B$, i.e. $\hat K_{\mathrm{eff}}(\alpha)\ge K/(1+B)$. As $\hat\sigma_\alpha^2>(\widehat{\mathrm{VaR}}_\alpha- \widehat{\operatorname{CVaR}}_\alpha)^2$, this gives $\alpha>1/(1+B)$.

    \item Keep the variance in eq. (\ref{eq:var_cvar}) constant at each optimizer call, as the  accuracy condition, which we do not study in this paper.
\end{itemize}
Applying these methods to CVaR leads to additional difficulties compared to Gibbs. Unlike $E_{z_0}$ in eq.~(\ref{eq:var_Gibbs}), $\mathrm{VaR}_\alpha$ does not cancel out and is replaced by its estimator from the same $K$ shots, which represents an approximation. We nevertheless provide a numerical result based on approximately solving eq.~(\ref{eq:Keff_cvar}) and keep further studies for future work.

\subsection{Practical aspects and concluding remarks}

In practice, $\hat\chi^2(Q_w \Vert P)$ is evaluated with eq.~(\ref{ea:chiestimator}) for a trial value of $\eta$, compared with $B$, and the hyper-parameter is updated iteratively until
eq.~(\ref{eq:rule}) holds.

The estimated "calibration function" deduced from eq.~(\ref{ea:chiestimator}), $\ln(1+\hat\chi^2(Q_\eta\Vert P))= \ln(K)-\ln(\hat K_{\mathrm{eff}}(\eta))$, is non-decreasing in $\eta$. Indeed, one can show that
\begin{equation}
    \frac{d}{d\eta}\ln\big(1+\hat\chi^2(Q_\eta\Vert P)\big) = 2\big(\bar E_\eta - \bar E_{2\eta}\big) \;\ge\; 0 ,
\end{equation}
where $\bar E_\eta=\sum_{k=1}^K E(k) e^{-\eta E(k)}/(\sum_{k=1}^K e^{-\eta E(k)})$ is the mean of the shot energies weighted by $e^{-\eta E}$, which decreases with
$\eta$, strictly as soon as two measured energies differ. The estimated "calibration function" thus rises from $0$ at
$\eta = 0$ to $\ln(K/m_0)$ (i.e. $\hat\chi^2$ rises to $K/m_0-1$) as $\eta \to \infty$, with $m_0$ the number of shots at the lowest measured energy. The rule in eq.~\eqref{eq:rule} therefore has a unique solution whenever $m_0 < K/(1+B)$, found by a one-dimensional numerical root search. Otherwise the state has already concentrated beyond the budget, and the rule returns a cap $\eta_{\max}$ (to avoid numerical divergence). The cap is a fixed number in the units of our rescaled Hamiltonians, $\eta_{\max}=500$ for $E_{z_0}=-N$. In practice one would stop the training once $m_0\ge K/(1+B)$, we continue it so that all methods use the same number of training shots.

The rule replaces the choice of $\eta$ by the choice of $B$, which is much easier to make. Indeed it works while being agnostic to the energy distribution of the problem. If the energies are changed into $aE+b$ with $a>0$, the solution of eq.~\eqref{eq:rule} becomes $\eta/a$ and the weights are unchanged, so solving eq.~\eqref{eq:rule} needs neither $E_{z_0}$ nor the energy scale of the problem, unlike an $\eta$ fixed a priori that acts as an oracle. 

In this work we use $B=1$, i.e.\ half of the shots are effective. 
We tried various values in numerical tests.
We observed that a smaller $B$ gives a more precise objective function but closer to the mean energy, i.e. a smaller tilt $\eta$. A larger $B$ gives a larger tilt $\eta$ but a noisier objective function. 


As we will observe numerically below, $\eta$ increases during the training for a fixed $B$. As the optimizer concentrates the state on low energies, the low-energy part of the distribution narrows, and the rule allows a larger $\eta$ for the same budget, which weights more lower energies. Conversely, if the distribution broadens, $\eta$ decreases and the estimate stays within the $\chi^2$-divergence shot budget. The rule does not by itself guarantee progress, it only keeps the tilt as sharp as the shots allow at every step. One could expect a positive feedback, where a narrower distribution allows a larger $\eta$, which concentrates the state further. This is precisely what we observe for the Gibbs objective function in Sec.~\ref{sec:constant-depth} and Fig.~\ref{fig:s25-ecdf} where $\eta$ increases during training as the concentration sharpens near the lowest energy.

\section{Numerical simulations}

The noiseless state-vector simulations were performed using QOKit~\cite{lykov2023fast}. We used QOKit solely as a classical QAOA simulation backend; the objective functions, the $\chi^2$-divergence rule, the shot-based estimators and the optimization procedures described below were implemented separately.

\subsection{Quadratic unconstrained combinatorial optimization problems}

In this section, $z_k\in\{0,1\}$ denotes the $k^{th}$ bit among the $N$ bits of one bitstring $z\in \{0,1\}^N$, unlike the $z_i\in \{0,1\}^N$ of the previous sections that denotes the $i^{th}$ bitstring sorted in energy among the $2^N$ possible bitstrings $\{z\}$. To lighten the notations, the use of $k,l$ or $i$ defines one case or the other.

We consider quadratic unconstrained combinatorial optimization (QUBO) problems \cite{Lucas_2014}, for which 
\begin{align}
E_z=\sum_{k,l=1}^N z_k Q_{kl} z_l,
    \label{eq:QUBO_0}
\end{align}
in eq. (\ref{eq:opt_pb}), with $Q_{kl} \in \mathbb{R}$ symmetric.

The problem Hamiltonian in eq. (\ref{eq:HP}) is more conveniently defined in terms of spin variables.
We first do the change of variable $s_k = (1-2 z_k)\in\{-1,1\}$ and obtain up to an irrelevant constant the problem Ising form \cite{Lucas_2014}
\begin{align}
E_z \rightarrow E_s= \sum_{k=1}^N h_k s_k + \sum_{k<l} J_{kl} s_k s_l.
    \label{eq:Ising}
\end{align}
Doing the substitution $s_k\rightarrow Z_k$, we deduce
\begin{align}
    H_P\rightarrow \sum_{k=1}^{N}h_{k}Z_k + \sum_{k<l}J_{kl}Z_k Z_l,
    \label{eq:HP_Ising}
\end{align}
where $Z_k$ is the $N$-qubit Pauli-Z operator, represented by a $2^N\times 2^N$ matrix in the computational basis.

We provide numerical evidence on the robustness of our QAOA with Gibbs scheme, in comparison with standard QAOA and CVaR QAOA with $\alpha$ set by the $\chi^2$-budget rule, on weighted Max-Cut and portfolio optimization problems.

\subsubsection{Weighted Max-Cut}

Max-Cut is the reference graph problem studied in the QAOA literature, see e.g. Refs. 	\cite{farhi2014qaoa,zhou2020qaoa}. We assign each node a bit $z_k \in \{0, 1\}$ to indicate which set it belongs to, and want to maximize the total weight $w_{kl}$ assigned to the edges that join the two sets. Considering the Ising form of Max-Cut, the QAOA problem Hamiltonian of eq.~(\ref{eq:HP_Ising}) becomes~\cite{zhou2020qaoa,blekos2024review}
\begin{align}
h_k=0, \qquad J_{kl}=\frac{w_{kl}}{2},
\end{align}
with $J_{kl}=0$ if nodes $k$ and $l$ are not connected (meaning null weights $w_{kl}$).

We decide to work with weighted graphs because unweighted ones are substantially easier to solve for QAOA, i.e. the QAOA output probability distribution usually concentrates easily on the optimal bitstring(s) \cite{zhou2020qaoa,yu2022adaptive}.
Furthermore, weighted graphs are usually closer to industrially useful applications of QAOA.

\begin{figure*}[t]
    \centering
    \includegraphics[width=0.7\linewidth]{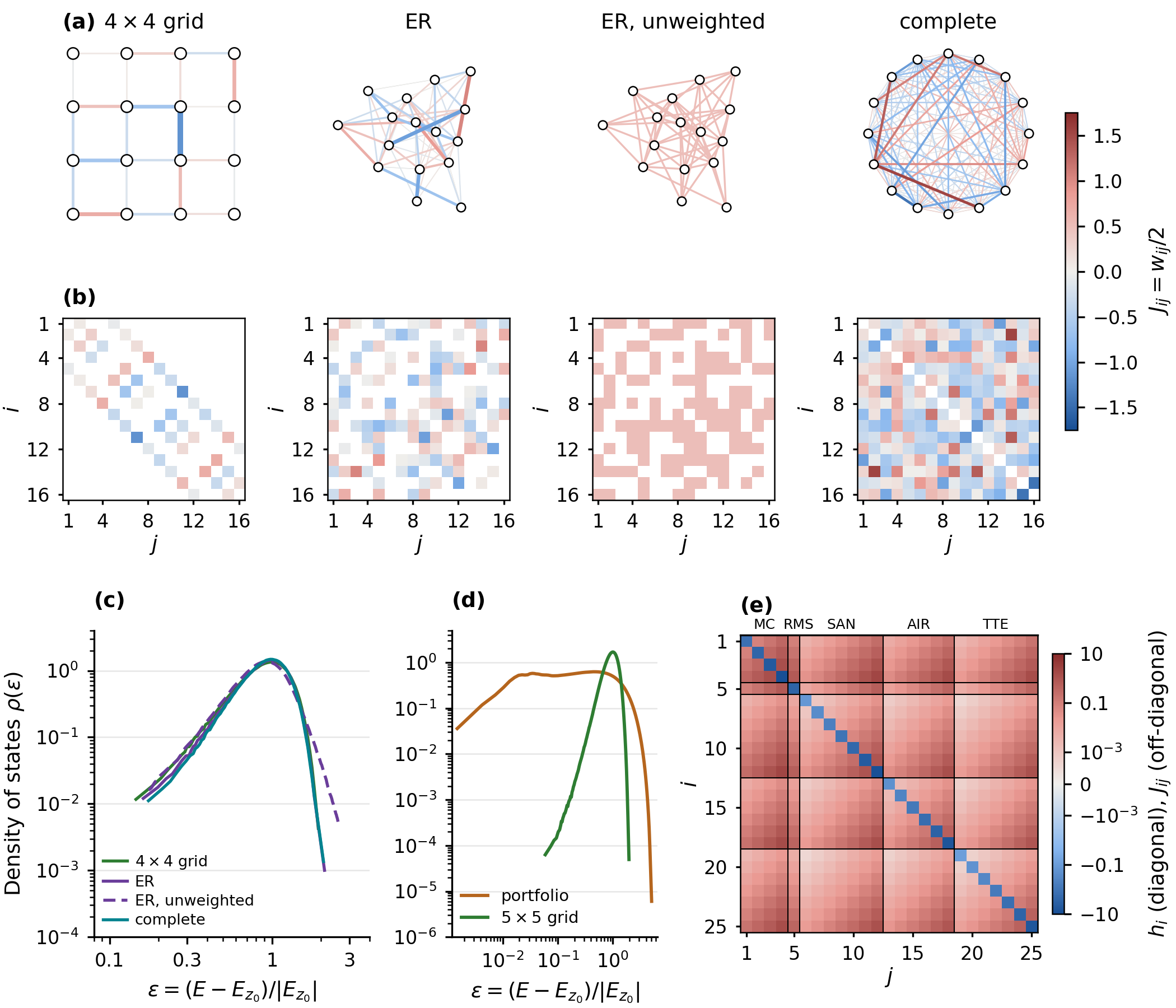}
    \caption{Problem instances and their density of states. (a) One example graph for each 16-qubit Max-Cut family and (b) its coupling matrix $J_{ij}=w_{ij}/2$.
    (c) Density of states $\rho(\varepsilon)$ of the four 16-qubit families, median over 100 instances each. (d) Density of states of the 25-qubit portfolio and grid instances, median over 32 instances each. (e) One portfolio optimization instance: $h_i$ on the diagonal and $J_{ij}$ off it, the lines separating the five assets. }
    \label{fig:maxcutexample}
\end{figure*}

We study 4 families of graphs:
\begin{itemize}
    \item  $4\times4$ (16 qubits) and $5\times5$ (25 qubits) grids with open boundaries and weights following a Gaussian distribution $\mathcal{N}(0,1)$.
    \item Erdős–Rényi (ER) at 16 qubits only, with probability 0.5 and weights following a Gaussian distribution $\mathcal{N}(0,1)$
    \item Unweighted Erdős–Rényi (ER) at 16 qubits only, with probability 0.5. 
    \item Complete graphs at 16 qubits only, with probability 0.5 and weights following a Gaussian distribution $\mathcal{N}(0,1)$
\end{itemize}
An example for each graph with their corresponding Ising matrix $J_{kl}$ is given in Fig.~\ref{fig:maxcutexample}.

\subsubsection{Portfolio optimization}

Our second testbed is an $n$ shares portfolio optimization case. 
The objective function to maximize represents a risk-return tradeoff, the risk being described using the  Markowitz form  \cite{markowitz1952portfolio,daniel2025portfolio}

\begin{align}
\label{eq:1stMar}
    \max_{0\le x\le x^{\max}}\ \sum_{i=1}^n \mu_i x_i - q\, \sum_{i,j=1}^n x_i C_{ij} x_j - \lambda(\sum_{i=1}^n x_i-1)^2,
\end{align}
with $\mu\in \mathbb{R}^n$ the expected return vector, $C\in \mathbb{R}^{n\times n}$ the covariance matrix between the different shares and $q\geq 0$ a so-called "risk-aversion weight". 
$x_i$ represents the fraction of a total monetary budget $M$ that must be bought for share $i\in\{1,...,n\}$, $x^{\max}\in]0,1]^n$ a cap on each asset fraction and $\lambda >0$ a Lagrange multiplier (or more exactly here a weight set a priori) associated with the constraint that the sum of the fractions $x_i$ must be as close as possible to $1$.

We consider the usual case where only a positive integer number of shares can be bought (no selling allowed), meaning that $x_i$ is constrained to take only discrete values in eq. (\ref{eq:1stMar}).
Considering the price  $X_i$ of share $i$, we can compute the maximum number
\begin{align}
\label{eq:zeta}
\zeta_i^{\max}=\max\left(1,\left\lfloor\frac{x_i^{\max}M}{X_i}\right\rfloor\right),
\end{align}
of asset $i$ that can be bought. The $\max(1,\cdot)$ keeps at least one bit per asset, so that all instances involve the same $n$
(on some instances it can make a share exceed its cap $x^{\max}$).
The allowed number of shares $\zeta_i\in\{0,\dots,\zeta_i^{\max}\}$ is then written in binary~\cite{Buonaiuto2023}, which requires a number of bits equal to
\begin{equation}
    N=\sum_{i=1}^n \left(\lfloor\log_2\zeta_i^{\max}\rfloor+1\right).
\end{equation}
Neither the cap nor the budget is a hard constraint, the bits of asset $i$ can represent more than $\zeta_i^{\max}$ shares and the quadratic penalty is soft and symmetric around the budget, so that the lowest energy bitstring can exceed both (Appendix~\ref{app:PO}). Then the problem can be turned into an Ising form and the $N$-qubit QAOA problem Hamiltonian of eq.~(\ref{eq:HP_Ising}) can be deduced,
\begin{align}
    h_k&=\pi_k\Big[\frac{\mu_{i_k}}{2}-\frac{q}{2}\sum_{l=1}^N C_{i_k i_l}\,\pi_l
        +\lambda\Big(1-\frac{1}{2}\sum_{l=1}^N\pi_l\Big)\Big], \nonumber\\
    J_{kl}&=\frac{\pi_k\pi_l}{2}\left(q\,C_{i_k i_l}+\lambda\right),
    \label{eq:hJ_portfolio}
\end{align}
where $i_k$ denotes the index of the asset associated with the $k^{th}$ binary variable in the encoding.
Appendix~\ref{app:PO}  and Ref.~\cite{Buonaiuto2023} give the steps to obtain this form as well as additional details.

All numerical results consider $n=5$ different shares described with $N=25$ qubits, and $q$ and $\lambda$ are scaled to balance the terms for each instance (Appendix~\ref{app:PO}).

The historical market data used to construct the portfolio instances are available from Yahoo Finance~\cite{yahoofinance}. The ticker symbols, sampling frequency, rolling time windows, and preprocessing applied to these data are specified in Appendix~\ref{app:PO}.

An Ising matrix $J$ and vector $h$ are shown in Fig.~\ref{fig:maxcutexample}(e). We observe patterns highlighted by the black lines, related to the binarization of each of the $5$ shares. Fig.~\ref{fig:maxcutexample}(c,d) shows the density of states $\rho(\varepsilon)$ of the Max-Cut and portfolio optimization instances, i.e. the fraction of the $2^N$ bitstrings per unit of residual energy $\varepsilon=(E_z-E_{z_0})/|E_{z_0}|$. The output probabilities of QAOA are approximately Boltzmann-distributed in energy~\cite{lotshaw2023boltzmann,DiezValle2026}, so that bitstrings of similar energy receive similar probabilities. Therefore the more states lie close to $E_{z_0}$, the more the probability is spread over them, and the smaller $P(z_0)$ is. At low depth the output probabilities are approximately Boltzmann-distributed in energy~\cite{lotshaw2023boltzmann,DiezValle2026}. The portfolio optimization problem class has a bigger energy spread and a much larger density of states near the lowest energy bitstring, which explains why it is harder to solve with QAOA than any of the weighted Max-Cut problems we consider. This will be further illustrated in sec.~\ref{sec:constant-depth}.

\subsection{Constant QAOA-depth}
\label{sec:constant-depth}

We first train QAOA at constant depth, $p \in \{8, 16, 32\}$ for many different instances of Max-Cut and portfolio optimization problems. 

For each instance, the angles
$(\boldsymbol{\gamma}, \boldsymbol{\beta})$ are initialized with a linear ramp whose end
points are drawn at random \cite{sack2021quantum}. The same initialization is used by every method.
The training is a sequence of BOBYQA calls \cite{powell2009bobyqa}, each warm-started from the parameters obtained by the previous result. 
Within a call, BOBYQA builds a quadratic model of the objective function within a trust region and uses it to propose trial angles, at which the true objective function is evaluated using $K$ shots.
The call ends when the trust radius reaches $\rho_{\mathrm{end}}$. All methods share the same total training shot budget $T$, fixed in advance (Table~\ref{tab:params}). Shots are drawn in batches of $K$, one batch for each evaluation of the objective function and, for the $\chi^2$-divergence based rule, one batch at the start of each call. $T$ is $K$ times the total number of batches of a run, so that every method draws the same number of batches. A batch is drawn only if it fits in the remaining shot budget. Otherwise the call in progress stops and returns its best point, and every run spends exactly $T$ shots.
 All computations are done using shots, apart from the reported metrics shown below that are computed exactly (state vector computation) from the final state. The algorithm is given in Alg.~\ref{alg:constant-depth}, and the values of all
parameters can be found in Appendix~\ref{app:cd-protocol}.

We compare the following methods:
\begin{itemize}
    \item S-QAOA, the standard QAOA, which uses the mean energy objective function.
    \item $\chi^2$-Gibbs-QAOA, which uses the Gibbs objective function with our $\chi^2$-divergence rule
    for a budget $B=1$. Here $\eta$ is computed at the start of every call. We draw $K$ shots at the
    current angles and find the value of $\eta$ for which the $\chi^2$-divergence of
    the Gibbs weights equals $B$, eqs. (\ref{eq:rule}) and (\ref{eq:Gibbs_solve}), by a Newton
    iteration on $\ln\eta$ safeguarded by bisection (Appendix~\ref{app:cd-protocol}).

    \item $\eta$-QAOA, which uses the Gibbs objective function with the constant
    $\eta=20/|E_{z_0}|$ of Ref.~\cite{li2020gibbs}, i.e. $\eta=1.25$ at 16 qubits and $\eta=0.8$ at
    25 qubits ($E_{z_0}=-N$ is the exact ground-state energy of our rescaled Hamiltonians, see Appendix ~\ref{app:PO}). It
    requires prior knowledge of $E_{z_0}$ and thus acts as an oracle.
    \item $\chi^2$-CVaR-QAOA, which uses the CVaR objective function of
    eq.~(\ref{eq:CVAR_estim}), where $\alpha$ is set at the start of every call by the $\chi^2$-divergence
    rule, with $B=1$, 5 and 10 ($K=10^3$ only). We draw $K$ shots at the current angles, sort their
    energies and try the energy levels in increasing order as $\widehat{\mathrm{VaR}}_\alpha$. We keep the first level for which $\hat K_{\mathrm{eff}}(\alpha)\ge K/(1+B)$, eq.~(\ref{eq:Keff_cvar}), and $\alpha K$ equals the number of shots below this level plus one.
\end{itemize}
We show the rank eCDF of Fig.~\ref{fig:ecdf} as a function of $k=i+1$, $\mathrm{eCDF}_r(k)\equiv\mathrm{eCDF}(E_{z_{k-1}})$, which represents the probability of sampling a bitstring at least as good as the $k$-th lowest-energy one (degenerate energy levels being being counted as a whole, leading to $\mathrm{eCDF}_r(1)=P_0$ using eq. (\ref{eq:P0})).
We also show the rank eCDF ratio to that of S-QAOA on the same instance, and medians over the instances.
\begin{algorithm}[H]
\caption{Constant-depth S-, $\eta$- and $\chi^2$-Gibbs-QAOA.}
\label{alg:constant-depth}
\begin{algorithmic}
\Require energies $E_z$, depth $p$, shots per evaluation $K$, shot budget $T$,
  method $\in\{\text{S-},\ \eta\text{-},\ \chi^2\text{-Gibbs-QAOA}\}$, $\eta_{\mathrm{fix}}$ or $B$, radii $\rho_{\mathrm{beg}},\rho_{\mathrm{end}}$, cap $\eta_{\max}$
\State $(\boldsymbol{\gamma}, \boldsymbol{\beta}) \gets$ linear ramp with random end points
\While{at least $K$ shots of the budget $T$ remain}
  \If{method $=\chi^2$-Gibbs-QAOA} \Comment{one $\eta$ per call}
    \State draw $z(1),\dots,z(K) \sim P(z,\boldsymbol{\gamma}, \boldsymbol{\beta})$, \ $E(k)=E_{z(k)}$
    \State $\eta \gets$ root of $\hat K_{\mathrm{eff}}(\eta)=K/(1+B)$ in $(0,\eta_{\max}]$
    \Statex \hspace{\algorithmicindent}\hspace{\algorithmicindent}(eq.~(\ref{eq:Gibbs_solve}); $\eta_{\max}$ if there is none)
  \ElsIf{method $=\eta$-QAOA}
    \State $\eta \gets \eta_{\mathrm{fix}}$
  \EndIf
  \If{method $=$ S-QAOA} \Comment{mean energy, eq.~(\ref{eq:mean_estimator})}
    \State $(\boldsymbol{\gamma}, \boldsymbol{\beta}) \gets \textsc{Bobyqa}\big(\widehat{\bra{\Psi}H_P\ket{\Psi}},\ (\boldsymbol{\gamma}, \boldsymbol{\beta}),\ \rho_{\mathrm{beg}}\!\to\!\rho_{\mathrm{end}}\big)$
  \Else \Comment{Gibbs, eq.~(\ref{eq:Gibbs_estimator})}
    \State $(\boldsymbol{\gamma}, \boldsymbol{\beta}) \gets \textsc{Bobyqa}\big(\hat F_\eta,\ (\boldsymbol{\gamma}, \boldsymbol{\beta}),\ \rho_{\mathrm{beg}}\!\to\!\rho_{\mathrm{end}}\big)$
  \EndIf
\EndWhile
\State \Return $(\boldsymbol{\gamma}, \boldsymbol{\beta})$
\end{algorithmic}
\end{algorithm}

\begin{figure*}[t!]
    \centering
    \includegraphics[width=0.75\linewidth]{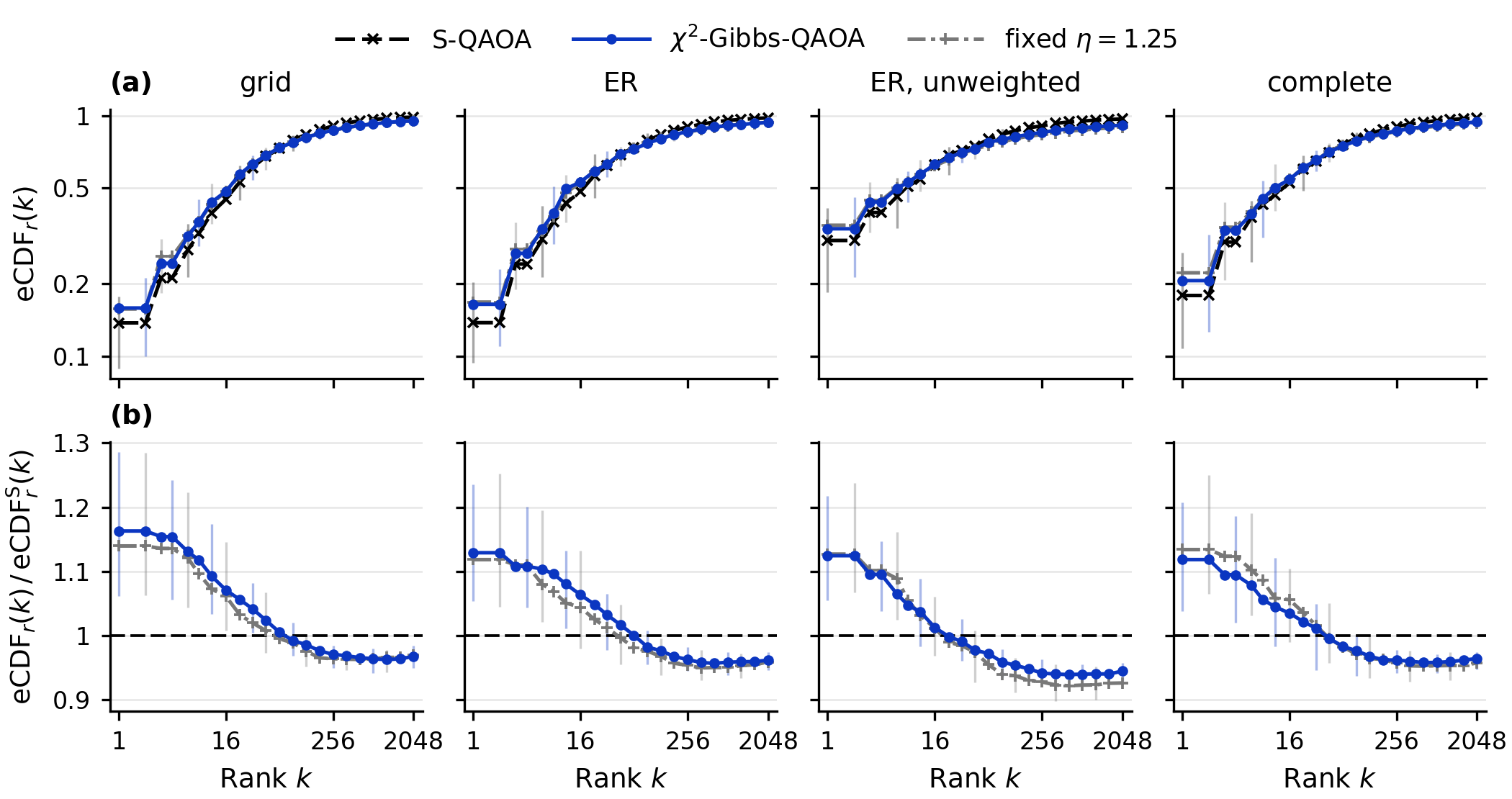}
    \caption{16-qubit ($2^{16}$ possible bitstrings) Max-Cut families at depth $p=8$, $K=10^3$ for 100 instances each. (a) $\mathrm{eCDF}_r(k)$, the probability of sampling a bitstring at least as good as the $k$-th lowest-energy one,
for $k \le 2048$. (b) eCDF$_r$ ratio to S-QAOA on the same instance (dashed line: S-QAOA).}
    \label{fig:s16-ecdf}
\end{figure*}

We first compare the 4 Max-Cut
graph types on 16-qubit graphs, with 100 instances randomly chosen for each type and $K=10^3$
(Fig.~\ref{fig:s16-ecdf} for $p=8$, Appendix~\ref{app:s16} for $p=16$ and 32). As expected from the
density of states given in Fig.~\ref{fig:maxcutexample}, the three weighted families share similar
$\mathrm{eCDF}$ profiles under the same training conditions, while the unweighted ER concentrates much more around the lowest energy bitstring. In fact, for $p=32$ it reaches a lowest-energy probability $P_0$ above 0.8
(Fig.~\ref{fig:s16-p32}). As a consequence, we drop the unweighted graphs from the larger 25-qubit study. Due to the similarity in the remaining graphs profiles, we only keep Max-Cut grid
family.

\begin{figure*}[t!]
    \centering
    \includegraphics[width=0.9\linewidth]{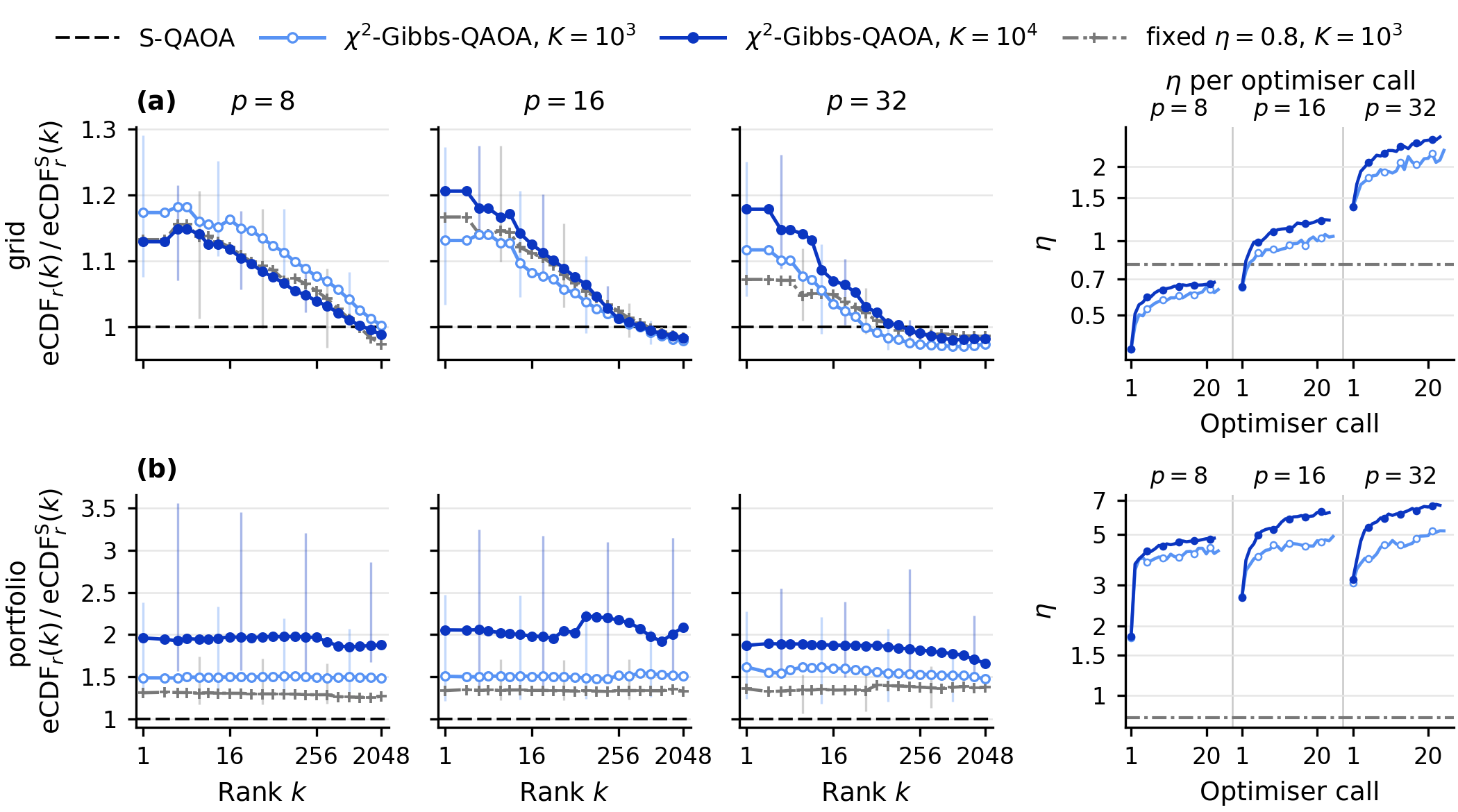}
    \caption{25-qubit ($2^{25}$ possible bitstrings) Max-Cut grid  (a) and portfolio optimization (b) at $p=8$, 16 and 32. Columns 1-3: median over the 32 instances of the ratio $\mathrm{eCDF}_r(k)/\mathrm{eCDF}_r^{\mathrm{S}}(k)$ to S-QAOA at the same $K$ on the same instance, with interquartile bars
    , for $\chi^2$-Gibbs-QAOA ($B=1$) at $K=10^3$ and $10^4$ and for $\eta$-QAOA with $\eta=0.8$ ($K=10^3$ only). Column 4: median over the instances of the $\eta$ chosen by the rule at each optimizer call (calls at the cap included); dash-dotted line: $\eta=0.8$.}
    \label{fig:s25-ecdf}
\end{figure*}

\begin{figure*}[t!]
    \centering
    \includegraphics[width=0.8\textwidth]{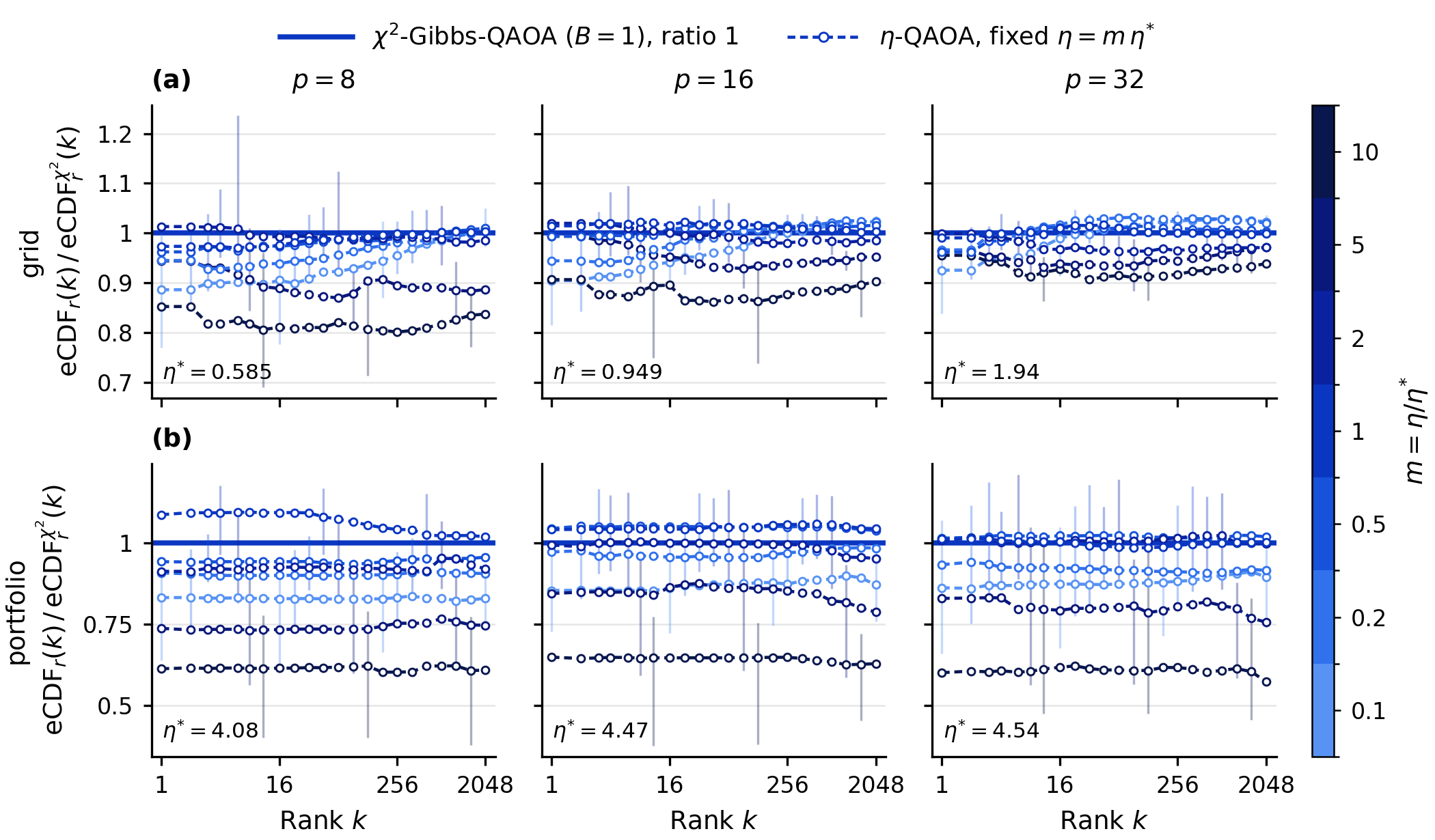}
    \caption{25 qubits instances ($2^{25}$ bitstrings) with fixed-$\eta$ sweep (exploratory), $K=10^3$: (a) Max-Cut grid and (b) portfolio, at $p=8$, 16 and 32. Median over the 32 instances of the ratio $\mathrm{eCDF}_r(k)/\mathrm{eCDF}_r^{\chi^2}(k)$ of $\eta$-QAOA with $\eta=m\,\eta^\star$ (dashed, color bar) to $\chi^2$-Gibbs-QAOA ($B=1$, solid line) on the same instance, with interquartile bars. The value of $\eta^\star$ is given in each panel.}
    \label{fig:eta-sweep}
\end{figure*}

\begin{figure*}[t!]
    \centering
    \includegraphics[width=\textwidth]{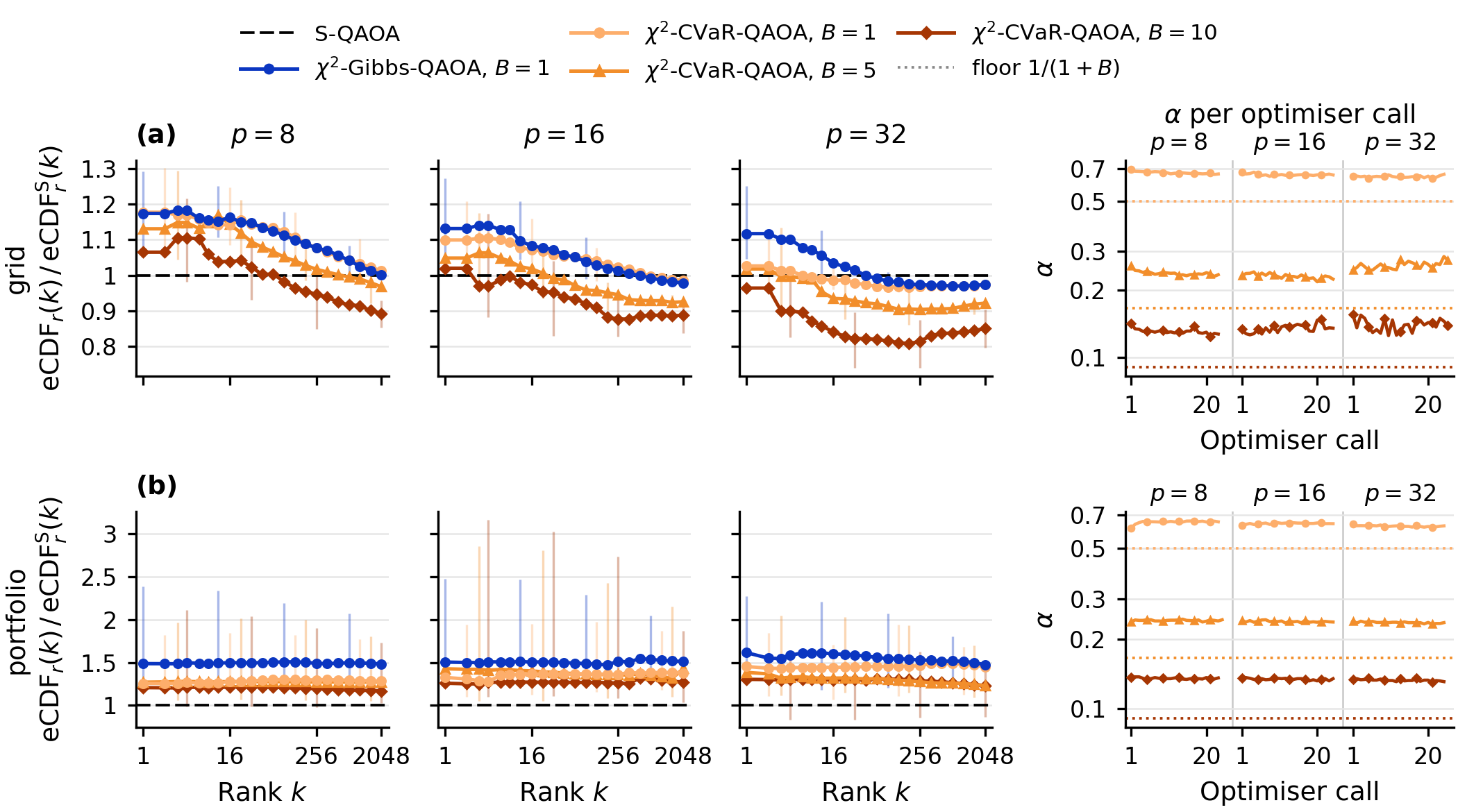}
    \caption{Same as Fig.~\ref{fig:s25-ecdf} for $\chi^2$-CVaR-QAOA ($B=1$, 5 and 10, exploratory)
     at $K=10^3$, with $\chi^2$-Gibbs-QAOA ($B=1$) repeated for comparison. Column 4: median over the instances of the $\alpha$ chosen by the rule at each optimizer call (calls at the cap included); dotted lines: lower bound $1/(1+B)$.}
    \label{fig:s25-cvar}
\end{figure*}

We now run S-QAOA, $\chi^2$-Gibbs-QAOA and $\eta$-QAOA on 32 Max-Cut grid instances and 32 portfolio instances, both at 25 qubits,
for $K=10^3$ and $K=10^4$ shots per evaluation of the objective function
($\eta$-QAOA at $K=10^3$ only). The results are shown in
Fig.~\ref{fig:s25-ecdf}.  For clarity we only show $B=1$, the budgets $B=0.1$, 3 and 10 were also run on the same instances and are shown in Appendix~\ref{app:s25-results} (Fig.~\ref{fig:app-ecdf-budgets}). $B=3$ is close to $B=1$, while $B=0.1$ and $B=10$ give a lower $P_0$.
First, we confirm that the portfolio optimization problem is much harder than the
Max-Cut grid problem. Indeed, the median probability $P_0$ of the lowest-energy bitstrings (two of them for Max-Cut, by $\mathbb{Z}_2$ symmetry) for S-QAOA at $K=10^4$ and $p=32$ is
$P_0 = 0.146$ for Max-Cut grid against $P_0 \sim 10^{-5}$ for the portfolio (which is still far above the uniform value for $2^{25}$ bitstrings, of order $10^{-8}$). The eCDFs are given in Appendix~\ref{app:s25-results} (Fig.~\ref{fig:app-ecdf-abs}).

For the Max-Cut grid family, $\chi^2$-Gibbs-QAOA increases the probability of the low-energy tail by a
factor 1.1 to 1.2. For $p=8$ and $p=16$, the scale of $\eta$ found by our $\chi^2$-divergence rule is
comparable to the oracle value of $\eta$-QAOA. At $p=32$, however, the typical $\eta$ chosen by the
rule is more than twice the oracle value $\eta=0.8$, and $\chi^2$-Gibbs-QAOA does better than $\eta$-QAOA
(Fig.~\ref{fig:s25-ecdf}).

On portfolio optimization, the gain is larger.
At $K=10^3$, the median over the instances of the ratio of $P_0$ to that of S-QAOA is 1.5 to 1.6, against 1.1 to 1.2 on Max-Cut grid, and the ratio to $\eta$-QAOA
is 1.10 to 1.15. 
Unlike for Max-Cut grid, this ratio stays nearly the same over the 2048 lowest energy bitstrings
(Fig.~\ref{fig:s25-ecdf}). 
At $K=10^4$, the ratio to S-QAOA reaches about 2. This gain comes with a
higher mean energy than S-QAOA as shown in Fig.~\ref{fig:mean-energy}, but this does not hamper the usage of QAOA as a sampler since we are only interested in the probabilities of low energy bitstring, even if that implies more probability mass on the high energy tail of the distribution as well.  In terms of eq.~(\ref{eq:concentration}), a ratio of 1.5 on the eCDF means about 30\% fewer sampling shots for the same probability of success $1-\delta$, both methods having used the same training shots.

To understand where the gain comes from, let us define $\eta^\star$, the typical $\eta$ chosen by the $\chi^2$-divergence based rule. It is computed in two steps from the $\chi^2$-Gibbs-QAOA runs at $K=10^3$. It is computed separately for each family and each depth $p$. First, for each instance, we take the median of the values of $\eta$ set at the successive BOBYQA calls of its run, leaving out the calls where $\eta$ reached its maximum cap value ($\eta = 500$). Then $\eta^\star$ is the median of these 32 values. There is thus one $\eta^\star$ per family and per depth, common to the 32 instances. We then run $\eta$-QAOA with seven fixed values $\eta=m\,\eta^\star$, from $m=0.1$
to 10, i.e.\ $\eta$ from 0.06 to 45 over the families and depths (Fig.~\ref{fig:eta-sweep}, eCDFs in Fig.~\ref{fig:app-ecdf-sweep}). Since $\eta^\star$ is computed on the same instances, the sweep favors $\eta$-QAOA. For $m$ between
0.5 and 2, the median ratio of $P_0$ to that of $\chi^2$-Gibbs-QAOA stays between 0.91 and 1.09, and no
value of $m$ does significantly better (Appendix~\ref{app:cd-protocol}). With $m=0.1$ and $m=10$,
$P_0$ drops by up to 17\% and 40\% respectively. The result therefore depends mainly on the order of
magnitude of $\eta$, which changes by a factor 8 between the families and depths, and which the rule
finds without knowing $E_{z_0}$. We conclude that the $\chi^2$-divergence rule is able to find automatically a value of
$\eta$ as good as the best fixed value we tested.
Note also that the value of $\eta$ chosen by the $\chi^2$-budget rule grows with $p$, by a factor 3.3 from $p=8$ to $32$ on Max-Cut grid.

Finally, we apply the $\chi^2$-budget rule to the CVaR objective function (Fig.~\ref{fig:s25-cvar}), with the cautions mentioned in Sec. \ref{sec:rule_Gibbs_CVaR}. For $B=1$,
the rule selects $\alpha\simeq0.65$ on both families and at every depth. On Max-Cut grid,
$\chi^2$-CVaR-QAOA follows $\chi^2$-Gibbs-QAOA at $p=8$ and 16, but fails at $p=32$: the median ratio of $P_0$ to that of S-QAOA is 1.03, against 1.12 for $\chi^2$-Gibbs-QAOA. On portfolio optimization, it keeps a gain over S-QAOA at every depth (1.26 to 1.45) but stays below $\chi^2$-Gibbs-QAOA
(1.5 to 1.6). A larger $B$ (smaller $K_{\rm eff}$) gives a smaller $\alpha$, about $\alpha=0.24$ for $B=5$ and $\alpha=0.13$ for $B=10$, and
does not help. On the grid the results get worse as $B$ grows, and on portfolio optimization the three values of $B$ give close results. In no case does $\chi^2$-CVaR-QAOA do significantly better than $\chi^2$-Gibbs-QAOA.  

Two effects explain this. First, unlike $\eta$, $\alpha$ hardly moves during the training
(Fig.~\ref{fig:s25-cvar}, column 4). The budget only bounds it from below, $\alpha>1/(1+B)$ by
eq.~(\ref{eq:chi2_cvar}), and above this bound $\alpha$ depends on the shape of the measured
distribution but not on its width, since the $\chi^2$-divergence of the CVaR weights is unchanged when the energies are rescaled. So when the distribution narrows with a larger depth $p$, $\eta$ grows while
$\alpha$ stays the same. Secondly, CVaR has a hard cutoff and only the $\lceil\alpha K\rceil$ lowest measured energies enter its estimator in eq.~(\ref{eq:CVAR_estim}). When at least that many shots are at the lowest energy $E_{z_0}$, the estimator equals $E_{z_0}$, its lowest possible value, whatever the other shots. This matters when the state has concentrated beyond the budget, $m_0\ge K/(1+B)$, with $m_0$ the number of shots of the rule at the lowest measured energy (which is $E_{z_0}$ in all these calls), the rule is then at its cap, as the Gibbs rule that returns $\eta_{\max}$ (Sec.~\ref{sec:chi2}). In that case, the second measured level already satisfies $\hat K_{\mathrm{eff}}=m_0\ge K/(1+B)$ in eq.~(\ref{eq:Keff_cvar}), and the rule gives $\alpha K=m_0+1$. The cutoff is thus placed just above the number of shots already at the lowest energy, and the objective function has little left to reward. On Max-Cut grid, BOBYQA then returns the angles it started from in about one call in four, the shots of the call being spent without changing the angles, against fewer than one call in ten when the rule is not at its cap. At $p=32$ on Max-Cut grid, the rule is at its cap in 6\%, 42\% and 76\% of the calls for $B=1$, 5 and 10, so that this second effect mostly concerns the larger budgets. The rule is never at its cap on portfolio optimization, where at most a few shots are at the lowest measured energy.

The Gibbs objective
function has no such cutoff: at its cap $\eta_{\max}$, it keeps rewarding concentration in agreement with our reasoning in Sec.~\ref{sec:chi2}.
This highlights the interest of this objective function as well as its alignment with the method we developed to automatically set the hyper-parameter.

\subsection{Iterative interpolation algorithm QAOA}
\label{sec:iterative_interpolation_experiment}

The iterative interpolation algorithm has been proposed in Ref.~\cite{apte2026iterative} (Algorithm~1 in that reference); our settings are given in Appendix~\ref{app:IIA}.
It takes advantage of the adiabatic inspiration of QAOA, as well as the similarity between parameters schedules with increasing depth, to make high depth circuit optimization (meaning hundreds of QAOA layers) feasible.
Our goal is to evaluate the impact of high depth on the eCDF and thus potentially on QAOA as a sampler performances.

The main idea of iterative interpolation is to represent the QAOA parameters with a small number of smooth basis functions, and to optimize these few coefficients rather than the $2p$ angles directly. First, a basis of smooth functions $g_j(t)$ on $[0,1]$, ordered in frequency, is introduced. The QAOA parameters are expanded in this basis, up to the $\mathcal{C}$ first terms 
\begin{align}
    &\gamma_k = \gamma (k/p) = \sum_{j=1}^\mathcal{C}u_j g_j(k/p) \nonumber \\
    &\beta_k = \beta (k/p) = \sum_{j=1}^\mathcal{C}v_j g_j(k/p). 
    \label{eq:ii_transform}
\end{align}
The optimization is then done on the new parameters $(\mathbf{u}, \mathbf{v})$ instead of $(\boldsymbol{\gamma},\boldsymbol{\beta})$. The main benefit comes from the fact that $\mathcal{C} < p$ such that the dimensionality of the search space  for the classical optimization is reduced. In doing so, and because the considered basis are ordered in frequency (like the discrete-Fourier or Chebyshev bases),
the method excludes all the $(\boldsymbol{\gamma},\boldsymbol{\beta} )$ schedules that would contain high frequency components from the search space, i.e. solutions that are far from the adiabatic regime.
$\mathcal{C}$ thus gives a control on how far the schedules can deviate from the adiabatic regime.
The algorithm starts at a small depth $p_{\min}$, and then iteratively increases $p$ all the way to $p_{\max}$ with a successive chain of warm starts obtained from the previous iteration results. More basis coefficients are added if the improvement of the objective function stalls.
Additional details can be found in Ref. \cite{apte2026iterative}.
 
\begin{figure*}[t!]
    \centering
    \includegraphics[width=0.9\textwidth]{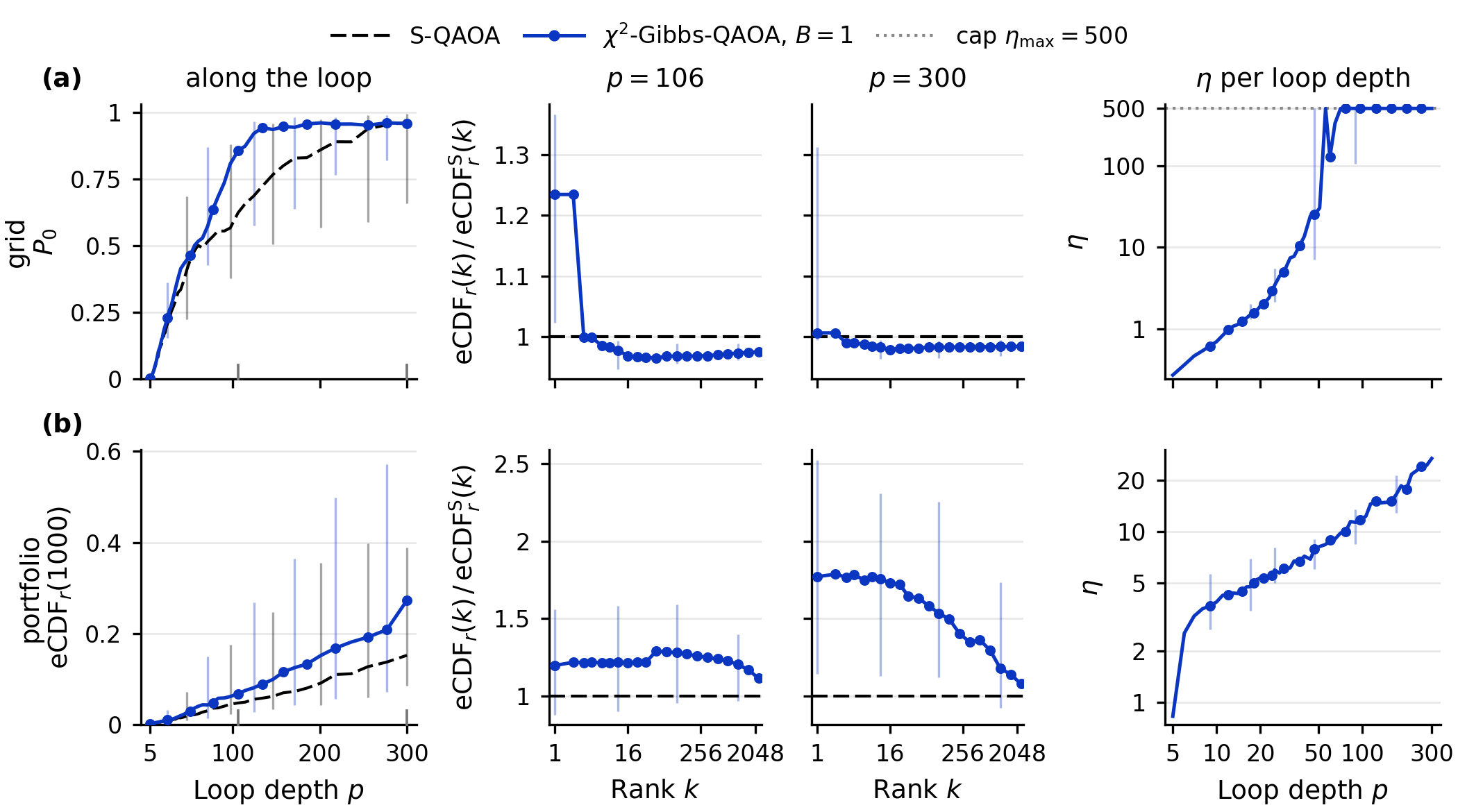}
    \caption{Iterative interpolation on the 25-qubit  ($2^{25}$ possible bitstrings) Max-Cut grid  (a) and portfolio optimization (b), from $p=5$ to 300 at $K=10^3$. Column 1: median over the 32 instances, with interquartile bars, of $P_0$ on Max-Cut grid and of $\mathrm{eCDF}_r(1000)$ on the portfolio at each depth, for S-QAOA and $\chi^2$-Gibbs-QAOA ($B=1$); $k=1000$ is interpolated between the stored ranks 885 and 1075; gray ticks: $p=106$ and 300. Columns 2-3: median over the instances of the ratio $\mathrm{eCDF}_r(k)/\mathrm{eCDF}_r^{\mathrm{S}}(k)$ to S-QAOA on the same instance, with interquartile bars, at $p=106$, chosen as the depth of the largest median ratio of $P_0$ on Max-Cut grid, and at $p=300$. Column 4: median over the instances of the $\eta$ chosen by the rule at each depth (calls at the cap included); dotted line: the cap $\eta_{\max}=500$.}
    \label{fig:dl}
\end{figure*}

We run the iterative interpolation algorithm with S-QAOA and $\chi^2$-Gibbs-QAOA ($B=1$) on the 32 Max-Cut grid and 32 portfolio optimization instances of Sec.~\ref{sec:constant-depth}, at $K=10^3$. The depth goes from $p_{\min}=5$ to $p_{\max}=300$ over 52 values, every $p$ up to 25 and then log-spaced, unlike in Ref. \cite{apte2026iterative} which uses a constant $\Delta p$ between steps. This choice is heuristic and is explained in Appendix~\ref{app:adiabatic}. For a given instance, both methods start at $p_{\min}$ from the same linear ramp as in Sec.~\ref{sec:constant-depth} and use the same random numbers to draw their shots, so that the comparison is paired. Each schedule is first expanded on $\mathcal{C}=4$ Chebyshev polynomials, and at every depth a single BOBYQA call optimizes the coefficients. For $\chi^2$-Gibbs-QAOA, $\eta$ is set once per depth, from $K$ shots drawn at the start of the call, with the cap $\eta_{\max}=500$. When the objective function improves by less than $10^{-3}$ in relative value from one depth to the next, one coefficient is added to each schedule, so that the two methods end with different numbers of coefficients. At $p=300$, a median of 24 per schedule for S-QAOA and 32.5 for $\chi^2$-Gibbs-QAOA on Max-Cut grid, and of 33.5 (S-QAOA) and 45 ($\chi^2$-Gibbs-QAOA) on portfolio optimization, the criterion being met more often with the Gibbs objective function. Unlike in Sec.~\ref{sec:constant-depth}, there is no {maximum} shot budget {T}, and a median run uses $1.0$ to $2.5\times10^{7}$ shots, those of the $\chi^2$-divergence based rule included.
As before, $P_0$ and $\mathrm{eCDF}_r(k)$ are computed exactly from the state reached at each depth. The settings of the optimizer and additional results are given in Appendix~\ref{app:dl}.

For both combinatorial problems, the probabilities associated to the low-energy bitstrings grow strongly with $p$ (Fig.~\ref{fig:dl}, column 1, which shows $P_0$ on Max-Cut grid and $\mathrm{eCDF}_r(1000)$ on portfolio optimization, for which the median $P_0$ stays of order $10^{-4}$). On Max-Cut grid, both reach a median $P_0$ of 0.96 at $p=300$ with about the same number of shots, but $\chi^2$-Gibbs-QAOA gets there at a smaller depth. At $p=106$, near $\chi^2$-Gibbs-QAOA's largest gain, its median $P_0$ is 0.86 against 0.62 for S-QAOA. The gain is on the lowest energy bitstring only, and at larger $k$ the ratio drops slightly below 1 (column 2). S-QAOA then catches up (column 3), and at $p=300$ the difference is not significant. S-QAOA still ends below $P_0=0.5$ on 5 instances, and $\chi^2$-Gibbs-QAOA on 2 of them.

On portfolio optimization, $\chi^2$-Gibbs-QAOA is better at every depth. The median ratio of its $\mathrm{eCDF}_r(1000)$ to that of S-QAOA stays between 1.1 and 1.4 along the loop. 
At $p=300$, the median ratio of $P_0$ is 1.8 and $\chi^2$-Gibbs-QAOA is ahead on 26 of the 32 instances, but the difference is not significant, as three of its runs drift away from the low energies during the loop. The ratio decreases with the rank but stays above 1 at every rank shown (column 3). The two median curves of column 1 are further apart than the median of the ratios, see Appendix~\ref{app:dl}.

$\chi^2$-Gibbs-QAOA uses more shots on portfolio optimization, about 1.4 times those of S-QAOA: it adds more coefficients, and its calls then need more evaluations. With the shots of a full S-QAOA run, it only reaches $p\simeq165$, where it is behind S-QAOA at $p=300$ on most instances.

On Max-Cut grid, $\eta$ reaches its cap around $p=56$ (column 4), and most instances stay there. This happens once the lowest energy bitstring holds about half of the shots of the rule, as expected for $B=1$ (Sec.~\ref{sec:chi2}). On portfolio optimization, where many bitstrings lie near the lowest energy bitstring (Fig.~\ref{fig:maxcutexample}-d), the lowest energy is typically drawn only once within the shots, and $\eta$ grows steadily, to about 27 at $p=300$.

The main advantage of the $\chi^2$-rule applied to the depth loop algorithm is that it reaches a given concentration at a lower depth, by a factor of up to about 2 on Max-Cut grid and 1.5 on portfolio optimization, which could matter in an intermediate regime before full FTQC, where the circuit depth is a limiting resource. The comparison is thus between the two procedures, each with the coefficients its rule adds; on the portfolio, $\chi^2$-Gibbs-QAOA also uses more shots.

\section{Conclusion}

In this work, we considered QAOA as a sampler under a finite shot budget, with the goal of assigning sufficient probability mass to bitstrings improving upon a reference solution. We related this operational goal to the eCDF, which provides a success criterion and an a posteriori benchmarking protocol for QAOA as a sampler. We justified the usage of the Gibbs objective function to favor concentration towards low energies, as a smooth objective function that maximizes an exponentially weighted average of the eCDF over energy thresholds.
However, the Gibbs objective function depends on a hyper-parameter $\eta$, crucial to control the "weight" placed on the low-energy tail.
Fixing automatically $\eta$ from sound theoretical foundations is highly desirable to ensure an efficient training of QAOA as a sampler and represented an open question. Connecting $\eta$ to a given QAOA shot budget $K$ represented another open question, important to control training accuracy. 

In this article, we proposed a method to resolve both problems.
Our main contribution is a $\chi^2$-divergence based rule that automatically sets $\eta$ at each optimizer call, from the shots, and can be related to controlling the effective sample size related to the Gibbs objective function, or the variance of the latter.
The rule does not require any prior knowledge of the lowest energy or of any energy scale.
We also applied the same principle to the CVaR hyper-parameter $\alpha$, although the 
resulting calibration is less direct because of the discrete hard cutoff of CVaR.

On the weighted Max-Cut and portfolio optimization instances studied here, $\chi^2$-Gibbs-QAOA increased the probability assigned to low-energy bitstrings compared with standard QAOA under equal training-shot budgets. A sweep over fixed values of $\eta$ showed that the rule automatically selected values comparable to the best fixed values tested, without instance-dependent calibration. Applying a similar effective sample size principle to CVaR led to smaller (or at most comparable) improvements in the reported experiments for $B=1$. Indeed, the hard cutoff of CVaR can make its sampled objective function locally flat, whereas the Gibbs objective function continues to distinguish probability redistributions within the low-energy tail.

The main advantage of applying the $\chi^2$-divergence based rule within the depth loop, using the iterative interpolation method, is that a given low-energy concentration is reached at a smaller QAOA depth. In the studied simulation cases, the required depth was reduced by factors of up to about $2$ for  Max-Cut grid and $1.5$ for portfolio optimization instances.
Such a reduction could be particularly relevant in an intermediate regime before full FTQC, where accessible circuit depth may remain a limiting resource.

Further work may investigate transfer learning, fundamental to a usage of QAOA in the FTQC regime, as well as variance-based calibration of CVaR. More generally, whether the eCDF concentration required for QAOA as a sampler can be reached with tractable shot, depth, and training costs remains an open question.

\section*{Acknowledgments}
We warmly thank K. Oublal for his illuminating and thorough review of this work, among others for the bound stated in eq. (\ref{eq:ecdf-gibbs2}).
We thank Y. Chatterjee for his contribution to the development of the model in Appendix \ref{app:PO}, and J.-P. Mascom\`ere and O. Ladhari for their insightful comments. We thank K. Gray, D. Cooper and H. Zong for providing and maintaining the CPU and GPU computing resources used for the simulations. We thank TotalEnergies for the permission to publish this work. The numerical simulations were performed using the open-source QOKit package developed by JPMorgan Chase and its contributors. The authors used a large language model as an assistant for proofreading and editing parts of the text.

\bibliographystyle{apsrev4-2}
\bibliography{bibliography}

\clearpage

\begin{appendix}

\section{Gibbs objective function as a mean of the eCDF}
\label{app:proof-ecdf}

We demonstrate here eq.~(\ref{eq:lemma_ecdf}).
Consider $e^{-\eta (E_z-E_{z_0})}=\int_{E_z-E_{z_0}}^{\infty}\eta e^{-\eta x}dx$, multiply by $P(z)$, sum over $z$ and obtain $\sum_{z=0}^{2^N-1} \int_{E_z-E_{z_0}}^\infty \eta e^{-\eta x}\,P(z)dx=\sum_{z=0}^{2^N-1} \int_0^\infty \eta e^{-\eta x}\,P(z)\mathbf{1}_{(E_z-E_{z_0}\le x)}dx$ as $E_z-E_{z_0}\ge 0$.
Exchange the sum and the integral, consider eq.~(\ref{eq:eCDF}) and obtain $\sum_{z=0}^{2^N-1} P(z)\,e^{-\eta (E_z-E_{z_0})}=\int_0^\infty\eta e^{-\eta x}\,\mathrm{eCDF}(E_{z_0}+x)dx$.
The shift by $E_{z_0}$ is what makes the weight a normalised density on $[0,\infty)$; without it the identity still holds but the weight has mass $e^{-\eta E_{z_0}}$.

\section{Portfolio optimization: from the Markowitz problem to the Ising form}
\label{app:PO}

\subsection{Markowitz problem with shares}
\label{sec:not}

Let $a_i^s$, $s\in\{0,\dots,S\}$, be the daily closing prices of asset $i\in\{1,\dots,n\}$ over the period that precedes the date at which the portfolio is built. The daily relative returns are $r_i^s=(a_i^s-a_i^{s-1})/a_i^{s-1}$, and the expected return and the covariance matrix are estimated by
\begin{align}
    \mu_i&=\frac{1}{S}\sum_{s=1}^S r_i^s, \nonumber\\
    C_{ij}&=\frac{1}{S-1}\sum_{s=1}^S(r_i^s-\mu_i)(r_j^s-\mu_j).
\end{align}
A portfolio is a vector $p=[p_1,\dots,p_n]^T$, where $x_i\ge 0$ is the fraction of the budget $M$ allocated to share $i$. The Markowitz problem is the return vs. risk trade-off optimization in eq. (\ref{eq:1stMar}).
We here consider that only a positive integer number $\zeta_i$ of shares can be bought, at the price $X_i=a_i^S$, so that
\begin{align}
    &x_i=\zeta_i\frac{X_i}{M},\quad \zeta_i\in\{0,\dots,\zeta_i^{\max}\},
\end{align}
where $\zeta_i^{\max}$ is given by eq. (\ref{eq:zeta}).

\subsection{Binary variables and QUBO form}

Following Ref.~\cite{Buonaiuto2023}, each $\zeta_i$ is written in binary on $N_i=\lfloor\log_2\zeta_i^{\max}\rfloor+1$ bits.
For each asset $i$, introduce binary variables $b_{i,m}\in\{0,1\}$ and write $\zeta_i=\sum_{m=0}^{N_i-1}2^m b_{i,m}$ and $x_i=\sum_{m=0}^{N_i-1}\frac{2^mX_i}{M}b_{i,m}$. We then enumerate all pairs $(i,m)$ by a single index $k=1,\ldots,N$, with $z_k=b_{i_k,m_k}$ and $\pi_k=2^{m_k}X_{i_k}/M$, which gives $N=\sum_{i=1}^n N_i$ bits $z_k\in\{0,1\}$ and
\begin{align}
    x_i=\sum_{k:\,i_k=i}\pi_k z_k.
    \label{eq:bin_pb}
\end{align}
Bit $k$ stands for $2^{m_k}$ shares of asset $i_k$, with $m_k\in\{0,\dots,N_{i_k}-1\}$, and is related to the fraction $\pi_k=2^{m_k}X_{i_k}/M$ of the monetary budget.
The $N_i$ bits can represent up to $2^{N_i}-1\ge\zeta_i^{\max}$ shares: values above $\zeta_i^{\max}$ are not removed by this encoding.
We obtain, after some computation, an equivalent form for the argument of eq. (\ref{eq:1stMar})
\begin{align}
    \mathcal{L}(z)
    =&\sum_{k}\mu_{i_k}\pi_k z_k
    - q\sum_{k,l}\pi_k C_{i_k i_l}\pi_l\, z_k z_l \nonumber\\
    &- \lambda\Big(\sum_{k}\pi_k z_k-1\Big)^2 .
\end{align}
Minimizing $E_z=-\mathcal{L}(z)$ is a QUBO problem. Using $z_k^2=z_k$, $E_z=\sum_{k,l} z_k Q_{kl} z_l+\lambda$ and
\begin{align}
    Q_{kl}=\pi_k\pi_l\left(q\,C_{i_k i_l}+\lambda\right)-\delta_{kl}\,\pi_k\left(\mu_{i_k}+2\lambda\right),
\end{align}
we recover the form in eq. (\ref{eq:QUBO_0}).
The penalty is soft and symmetric and does not forbid portfolios that exceed the budget.

\subsection{Ising form}

With the spin variables $s_k=1-2z_k\in\{1,-1\}$ and $s_k^2=1$, we obtain
\begin{align}
    E_s=\sum_{k} h_k s_k+\sum_{k<l}J_{kl}\,s_k s_l+\text{const},
\end{align}
with $h_k=-\frac{1}{2}\sum_{l}Q_{kl}$ and $J_{kl}=Q_{kl}/2$, which gives the Ising form in eq.~(\ref{eq:hJ_portfolio})
where the constant is dropped.

The weights $q$ and $\lambda$ are set independently for each instance from two dimensionless parameters $q_0$ and $\lambda_0$, whose values are typically of order one and are therefore easier to choose. We calibrate these weights by requiring that, at the midpoint $\zeta_i=\zeta_i^{\max}/2$ of the allowed number of shares, the $\ell_1$ norms of the gradients of the weighted risk and penalty terms are respectively $q_0$ and $\lambda_0$ times the $\ell_1$ norm of the gradient of the return term. This gives
\begin{align}
q &= q_0 \frac{ \displaystyle \sum_i \left|\mu_i \frac{X_i}{M}\right| }{ \displaystyle \sum_i \left| 2\frac{X_i}{M} \sum_j C_{ij}\frac{X_j}{M}\frac{\zeta_j^{\max}}{2} \right| },
\nonumber\\
\lambda &= \lambda_0 \frac{ \displaystyle \sum_i \left|\mu_i \frac{X_i}{M}\right| }{ \displaystyle \sum_i \left| 2\frac{X_i}{M} \left( \sum_j \frac{X_j}{M}\frac{\zeta_j^{\max}}{2}-1 \right) \right| }.
\end{align}

Finally, $h$ and $J$ are divided by a common factor so that the ground-state energy is $E_{z_0}=-N$.

\subsection{Instances}
\label{app:PO-instances}

We use $n=5$ shares of Euronext Paris (tickers MC, RMS, SAN, AIR and TTE), with $M=6000$ euros, $x_i^{\max}=1$ except 0.3 for RMS, $q_0=2$ and $\lambda_0=20$. The prices are the daily closing prices of Yahoo Finance~\cite{yahoofinance}, adjusted for dividends and splits, as downloaded on 23 October 2026, over the 180 calendar days that precede the date. The same adjusted prices are used for $X_i$; as they are revised at each dividend, $X_i$, and for some dates the number of bits, depend on the download date. Among the 5th of each month from November 2020 to August 2026, we keep the 32 dates for which the encoding takes $N=25$ qubits; they run from December 2022 to June 2026. 

Two dates less than six months apart share part of their 180-day windows, so the instances are not independent. For the  portfolio optimization tests, (Appendix~\ref{app:cd-protocol}), 
the variance of the mean over the instances of $D_i=\operatorname{logit}P_0^{(1)}(i)-\operatorname{logit}P_0^{(2)}(i)$, for two methods 1 and 2, is therefore a heteroskedasticity and autocorrelation consistent (HAC) estimate~\cite{NeweyWest1987},
\begin{align}
    V=\frac{1}{n_I(n_I-1)}\sum_{i,j=1}^{n_I} \omega(d_{ij})\,(D_i-\bar D)(D_j-\bar D),
\end{align}
with $n_I=32$ the number of instances, $\bar D$ the mean of the $D_i$, $d_{ij}$ the number of months between the dates of instances $i$ and $j$, and the Bartlett weights $\omega(d)=\max(0,1-d/6)$. $V$ is not taken smaller than its value for independent instances, and the interval uses a Student distribution with $n_I-1$ degrees of freedom.

\providecommand{\appcell}[2]{\parbox[t]{#1}{\raggedright\strut#2\strut}}

\section{Details of the constant-depth study}
\label{app:parameters}

\subsection{Parameters}
\label{app:cd-protocol}

The parameters of the constant-depth runs are given in Table~\ref{tab:params}. For a given instance, all the methods start from the same linear ramp and use the same random numbers to draw their shots, so that the comparisons are paired. Note that we compare the methods at equal shot budget, and not at convergence. A difference is called significant when the 95\% interval of the geometric mean over the instances of the odds ratio of $P_0$ excludes 1. On Max-Cut grid we use bootstrap-$t$ intervals~\cite{EfronTibshirani1993}, and on portfolio optimization a HAC variance~\cite{NeweyWest1987} (Appendix~\ref{app:PO-instances}), since dates less than six months apart share part of their 180-day windows.  At $p=32$, the tests fixed before the runs give an odds ratio of $P_0$ of $\chi^2$-Gibbs-QAOA to S-QAOA of 1.20 [1.13, 1.28] on Max-Cut grid and 1.76 [1.46, 2.12] on the portfolio, and to $\eta$-QAOA ($\eta=0.8$) of 1.07 [1.03, 1.10] and 1.30 [1.07, 1.59], with a Holm correction~\cite{Holm1979} within each family. The intervals of the fixed-$\eta$ sweep and of CVaR are exploratory and not corrected for multiplicity; for none of the 42 fixed values of $\eta$ does the interval of the odds ratio to $\chi^2$-Gibbs-QAOA lie above 1.

\begin{table*}[tbp]
\caption{Parameters of the constant-depth runs.}
\label{tab:params}
\footnotesize
\begin{ruledtabular}
\begin{tabular}{ll}
 & \appcell{0.80\textwidth}{} \\ \hline
Grid & \appcell{0.80\textwidth}{$5\times5$ open grid (40 edges), weights $w_{ij}$$\sim\mathcal{N}(0,1)$, 32 instances.} \\
Portfolio & \appcell{0.80\textwidth}{Eq.~(\ref{eq:1stMar}) on five shares, budget of 6000, $q_0=2$, $\lambda_0=20$ (Appendix~\ref{app:PO}), returns over 180 days, 32 monthly dates ($N=25$).} \\
16 qubits & \appcell{0.80\textwidth}{$4\times4$ open grid, Erd\H{o}s--R\'enyi graphs with edge probability 0.5 and complete graphs, weights $\mathcal{N}(0,1)$, and the same Erd\H{o}s--R\'enyi graphs with unit weights; 100 instances each, $K=10^3$, $\eta=1.25$ for $\eta$-QAOA, other parameters as at 25 qubits.} \\
Energy scale & \appcell{0.80\textwidth}{$E_{z_0}=-N$, from the exact enumeration of the $2^N$ energies.} \\
Depths & \appcell{0.80\textwidth}{$p=8$, 16 and 32, with the $2p$ angles of Eq.~(\ref{eq:ansatz}) optimized directly and $H_0=-\sum_iX_i$.} \\
Start & \appcell{0.80\textwidth}{Linear ramp from $\gamma_1=0$ to $\gamma_p\sim U(0.3,0.7)$ and from $\beta_1\sim U(0.4,1)$ to $\beta_p=0$, one per instance.} \\
Shot budget & \appcell{0.80\textwidth}{$T=212.5\,pK$, i.e.\ 1.7, 3.4 and $6.8\times10^6$ shots for $K=10^3$, the shots of the rule included.} \\
Optimizer & \appcell{0.80\textwidth}{BOBYQA (NLopt~\cite{NLopt,powell2009bobyqa}), $\rho_{\mathrm{beg}}=0.1$, $\rho_{\mathrm{end}}=10^{-2}$, relative tolerance $10^{-8}$, 22 to 47 calls per run (22 to 29 for S-, $\eta$- and $\chi^2$-Gibbs-QAOA at 25 qubits)
} \\
$\chi^2$ rule & \appcell{0.80\textwidth}{$B=1$, $K$ fresh shots at the start of each call, root of Eq.~(\ref{eq:rule}) in $(0,500]$, $\eta_{\max}=500$.} \\
Fixed $\eta$ & \appcell{0.80\textwidth}{$\eta=0.8$, and $\eta=m\eta^\star$ with $m=0.1$, 0.2, 0.5, 1, 2, 5 and 10, $\eta^\star=0.585$, 0.949 and 1.94 (grid) and 4.08, 4.47 and 4.54 (portfolio) at $p=8$, 16 and 32.} \\
$K=10^4$ & \appcell{0.80\textwidth}{S-QAOA and $\chi^2$-Gibbs-QAOA only.} \\
$\chi^2$-CVaR & \appcell{0.80\textwidth}{$B=1$, 5 and 10, $K$ fresh shots at the start of each call, smallest $\alpha$ within the budget; $K=10^3$ only.} \\
\end{tabular}
\end{ruledtabular}
\end{table*}

\subsection{eCDF at 25 qubits}
\label{app:s25-results}

\begin{figure*}[t!]
\centering
\includegraphics[width=0.85\textwidth]{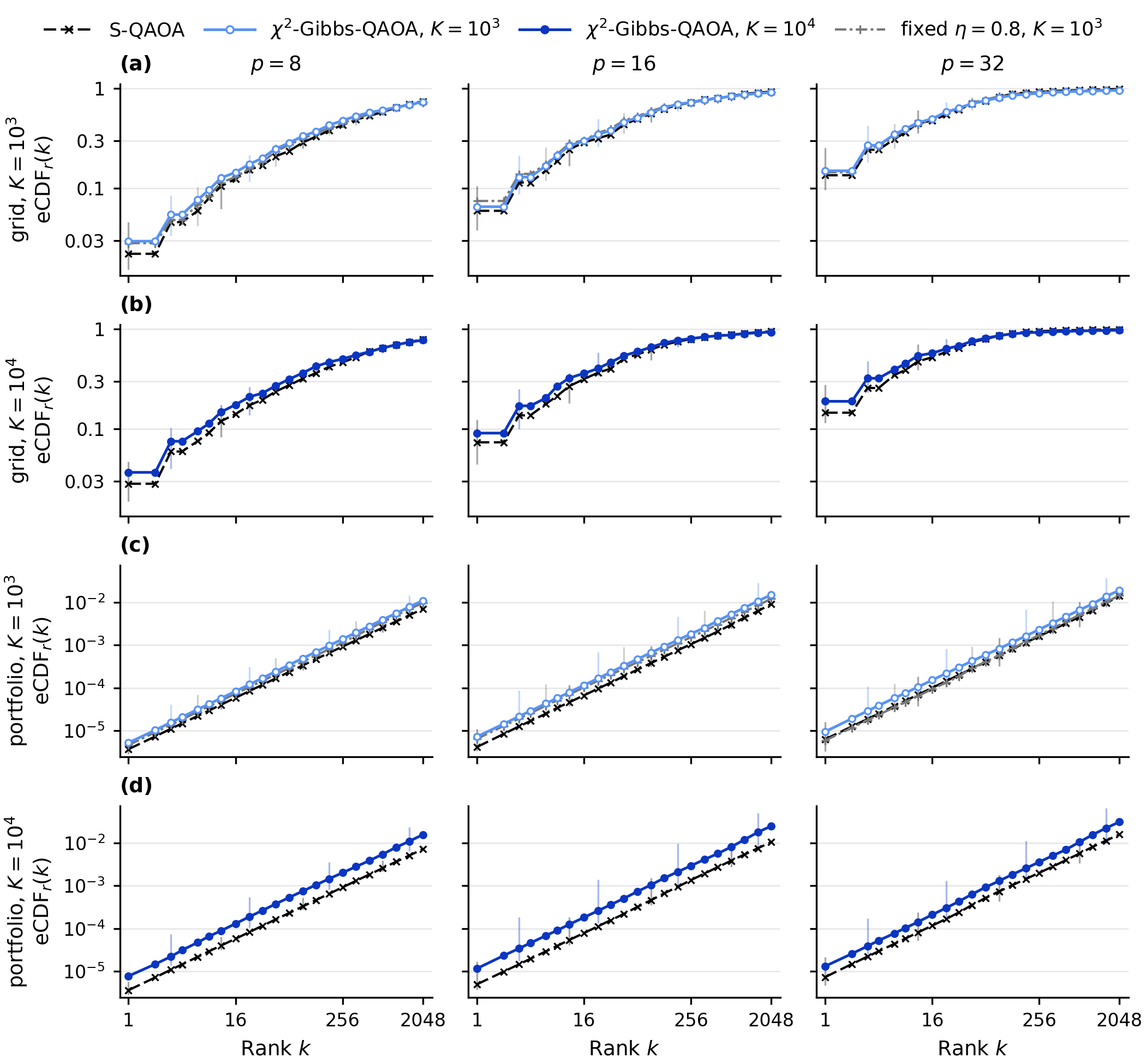}
\caption{$\mathrm{eCDF}_r(k)$ at 25 qubits, behind the ratios of Fig.~\ref{fig:s25-ecdf}: (a) grid, $K=10^3$; (b) Max-Cut grid, $K=10^4$; (c) portfolio optimization , $K=10^3$; (d) portfolio optimization, $K=10^4$, at $p=8$, 16 and 32. Medians over the 32 instances with interquartile bars.}
\label{fig:app-ecdf-abs}
\end{figure*}

\begin{figure*}[t!]
\centering
\includegraphics[width=0.9\textwidth]{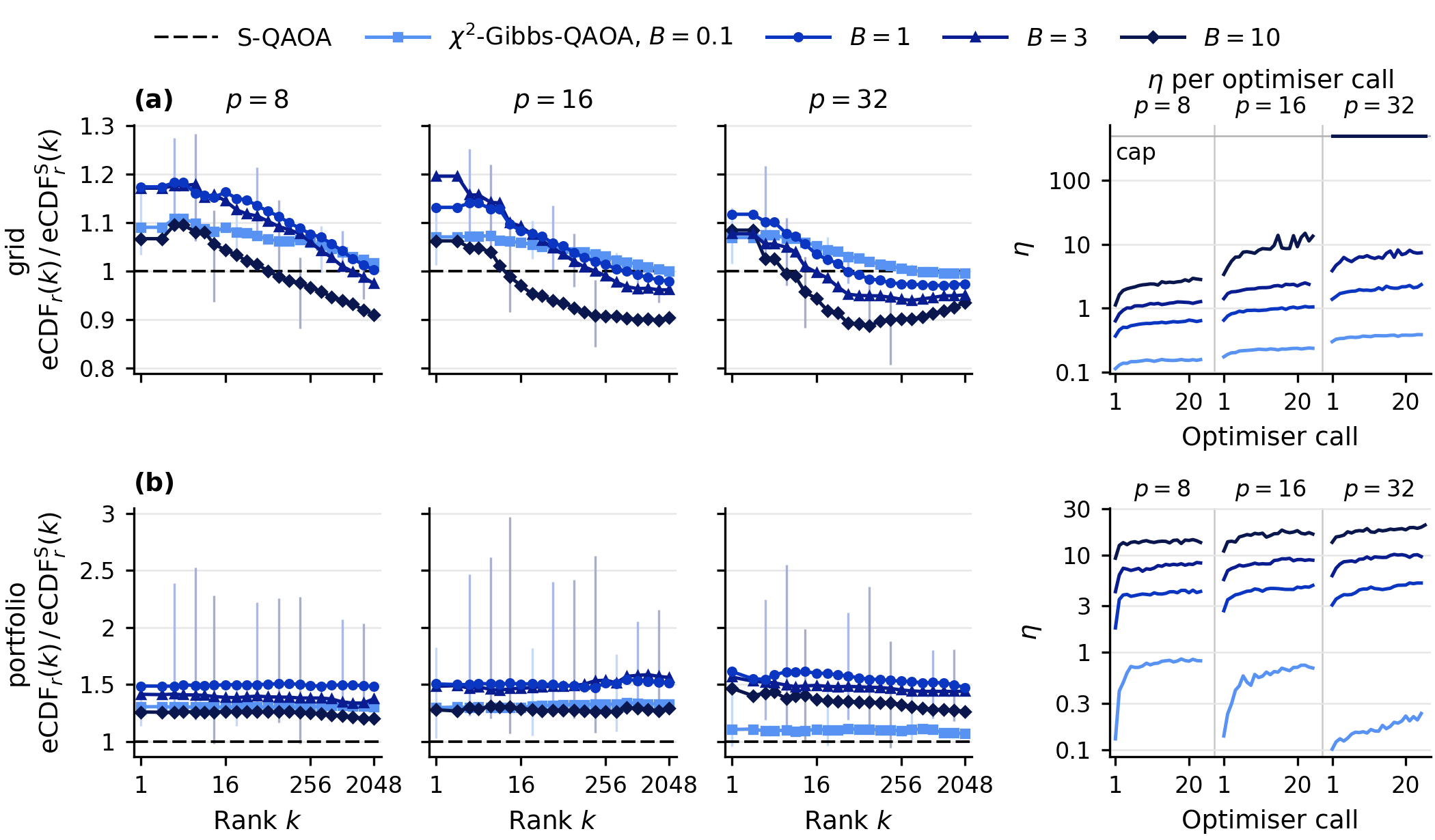}
\caption{Same as Fig.~\ref{fig:s25-ecdf} for $\chi^2$-Gibbs-QAOA with $B=0.1$, 1, 3 and 10 at $K=10^3$ (exploratory). The interquartile bars are shown for one budget in turn. Column 4: median over the instances of the $\eta$ chosen by the rule at each optimizer call (calls at the cap included).}
\label{fig:app-ecdf-budgets}
\end{figure*}

\begin{figure*}[t!]
\centering
\includegraphics[width=0.8\textwidth]{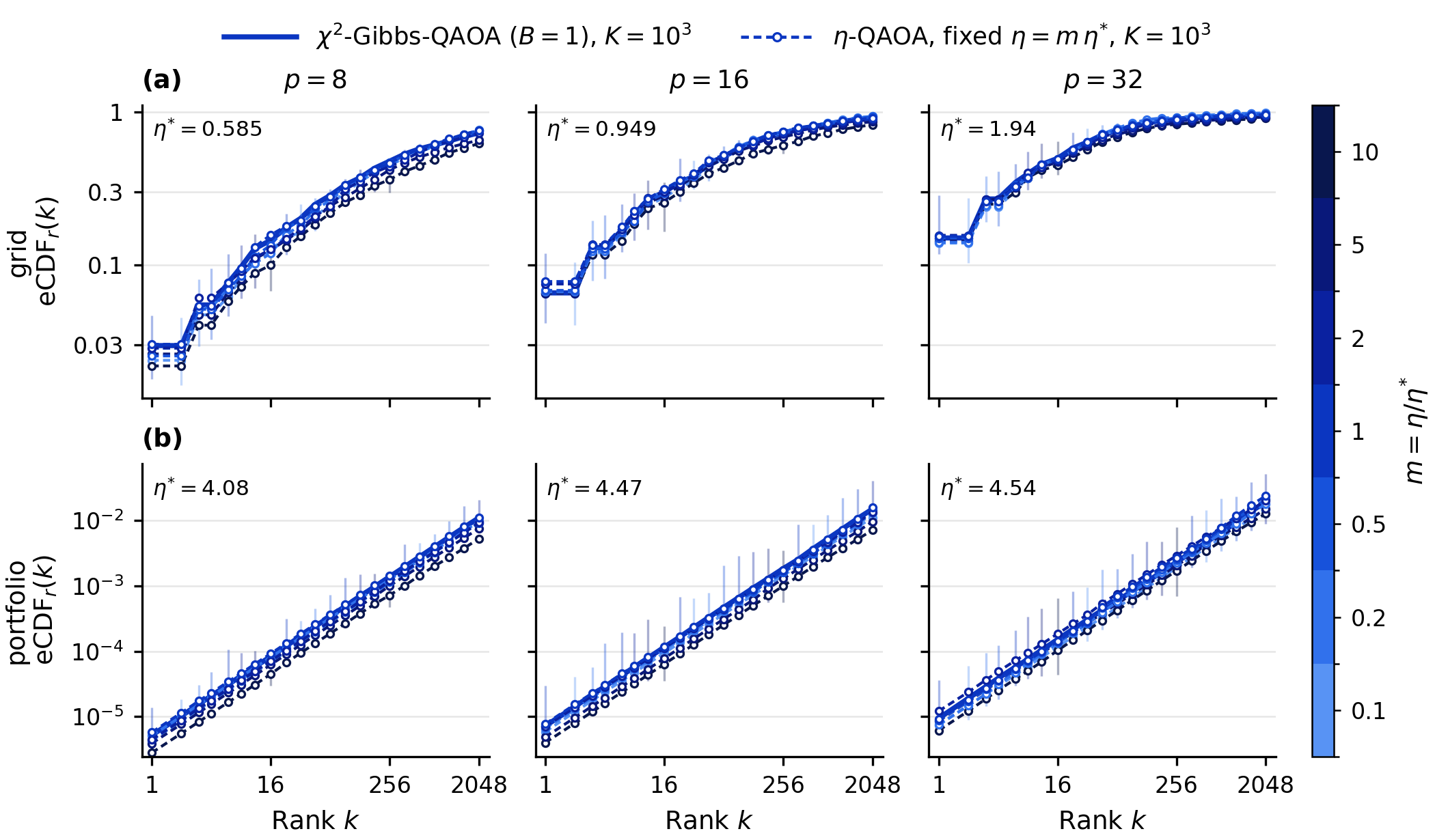}
\caption{$\mathrm{eCDF}_r(k)$ behind the ratios of Fig.~\ref{fig:eta-sweep}, $K=10^3$: (a) Max-Cut grid and (b) portfolio optimization at $p=8$, 16 and 32, for $\chi^2$-Gibbs-QAOA ($B=1$, solid line) and $\eta$-QAOA with $\eta=m\,\eta^\star$ (dashed, color bar). Medians over the 32 instances with interquartile bars.}
\label{fig:app-ecdf-sweep}
\end{figure*}

\begin{figure*}[t!]
\centering
\includegraphics[width=0.9\textwidth]{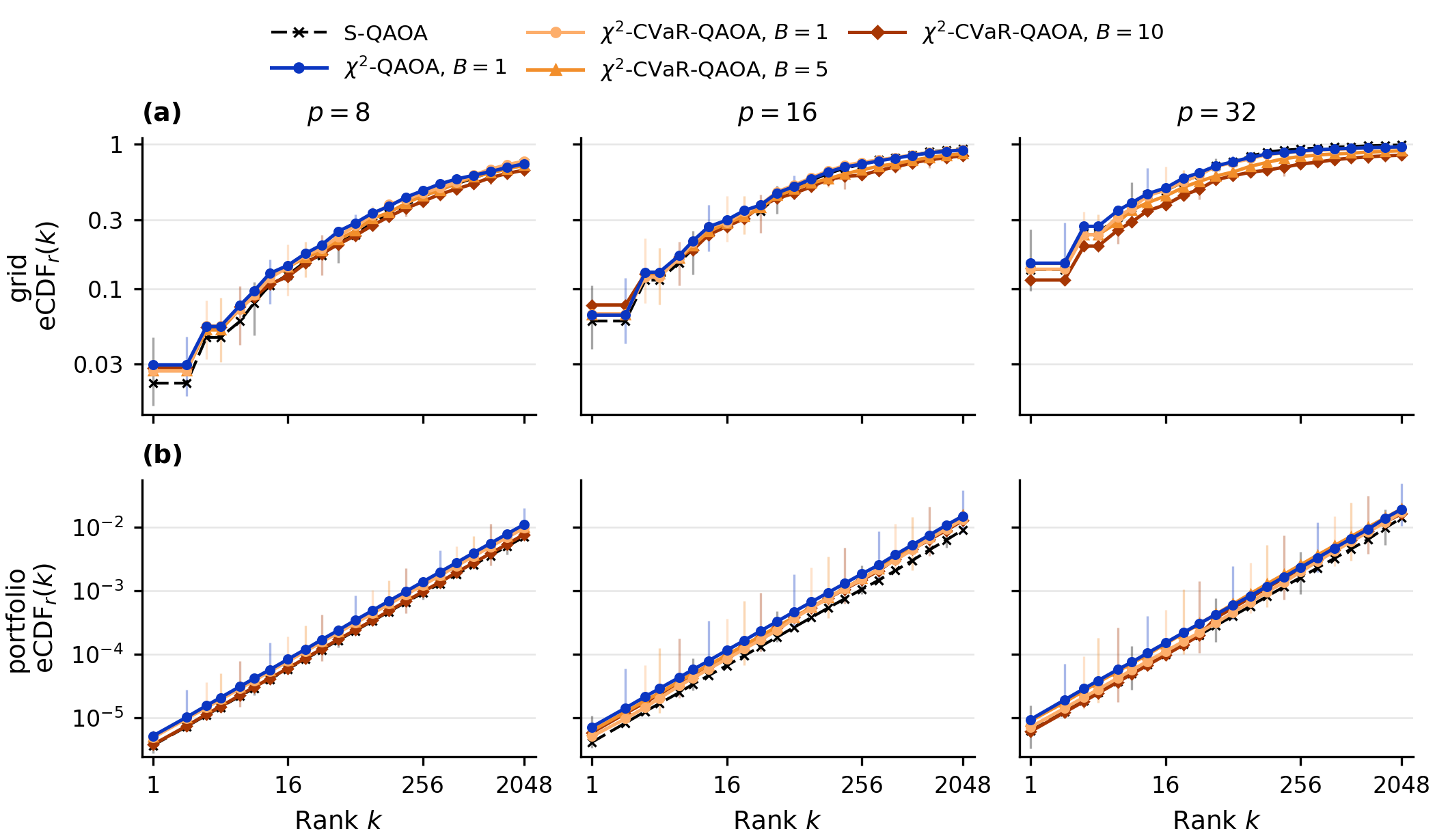}
\caption{$\mathrm{eCDF}_r(k)$ behind the ratios of Fig.~\ref{fig:s25-cvar}, $K=10^3$: (a) Max-Cut grid and (b) portfolio optimization  at $p=8$, 16 and 32, for S-QAOA, $\chi^2$-Gibbs-QAOA ($B=1$) and $\chi^2$-CVaR-QAOA ($B=1$, 5 and 10). Medians over the 32 instances at 25 qubits with interquartile bars.}
\label{fig:app-ecdf-cvar}
\end{figure*}

Figures~\ref{fig:app-ecdf-abs} and~\ref{fig:app-ecdf-sweep} give the $\mathrm{eCDF}_r(k)$ behind the ratios of Figs.~\ref{fig:s25-ecdf} and~\ref{fig:eta-sweep}. The ratios shown in the main text are medians over the instances of the ratio on each instance, which differ from the ratios of the medians shown here. On Max-Cut grid, the curves of the different methods join at large $k$, where the ratio goes back to 1. On the portfolio, they stay nearly parallel in logarithmic scale over the 2048 lowest states, consistent with the flat ratio of Fig.~\ref{fig:s25-ecdf}.  Figure~\ref{fig:app-ecdf-budgets} gives the ratios of Fig.~\ref{fig:s25-ecdf} for $\chi^2$-Gibbs-QAOA with $B=0.1$, 1, 3 and 10 at $K=10^3$. On Max-Cut grid at $p=32$, the rule is at its cap in 5\%, 26\% and 76\% of the calls for $B=1$, 3 and 10, the lowest energy level then holding more than $K/(1+B)$ of the shots. It is never at its cap on the portfolio.
Figure~\ref{fig:app-ecdf-cvar} gives the same curves for $\chi^2$-CVaR-QAOA.

\subsection{16-qubit study}
\label{app:s16}

\begin{figure*}[t!]
\centering
\includegraphics[width=\textwidth]{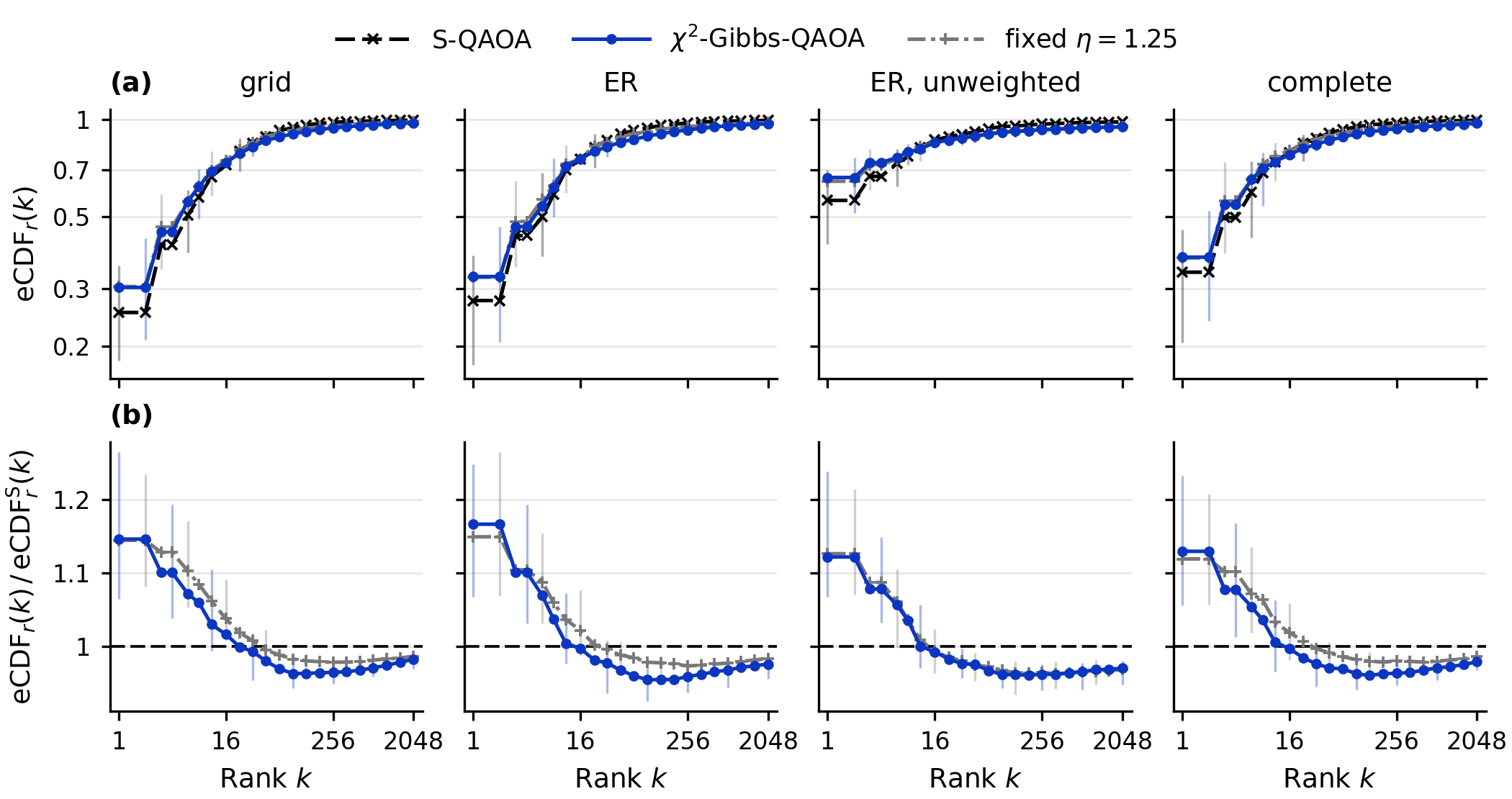}
\caption{Same as Fig.~\ref{fig:s16-ecdf} at $p=16$.}
\label{fig:s16-p16}
\end{figure*}

\begin{figure*}[t!]
\centering
\includegraphics[width=\textwidth]{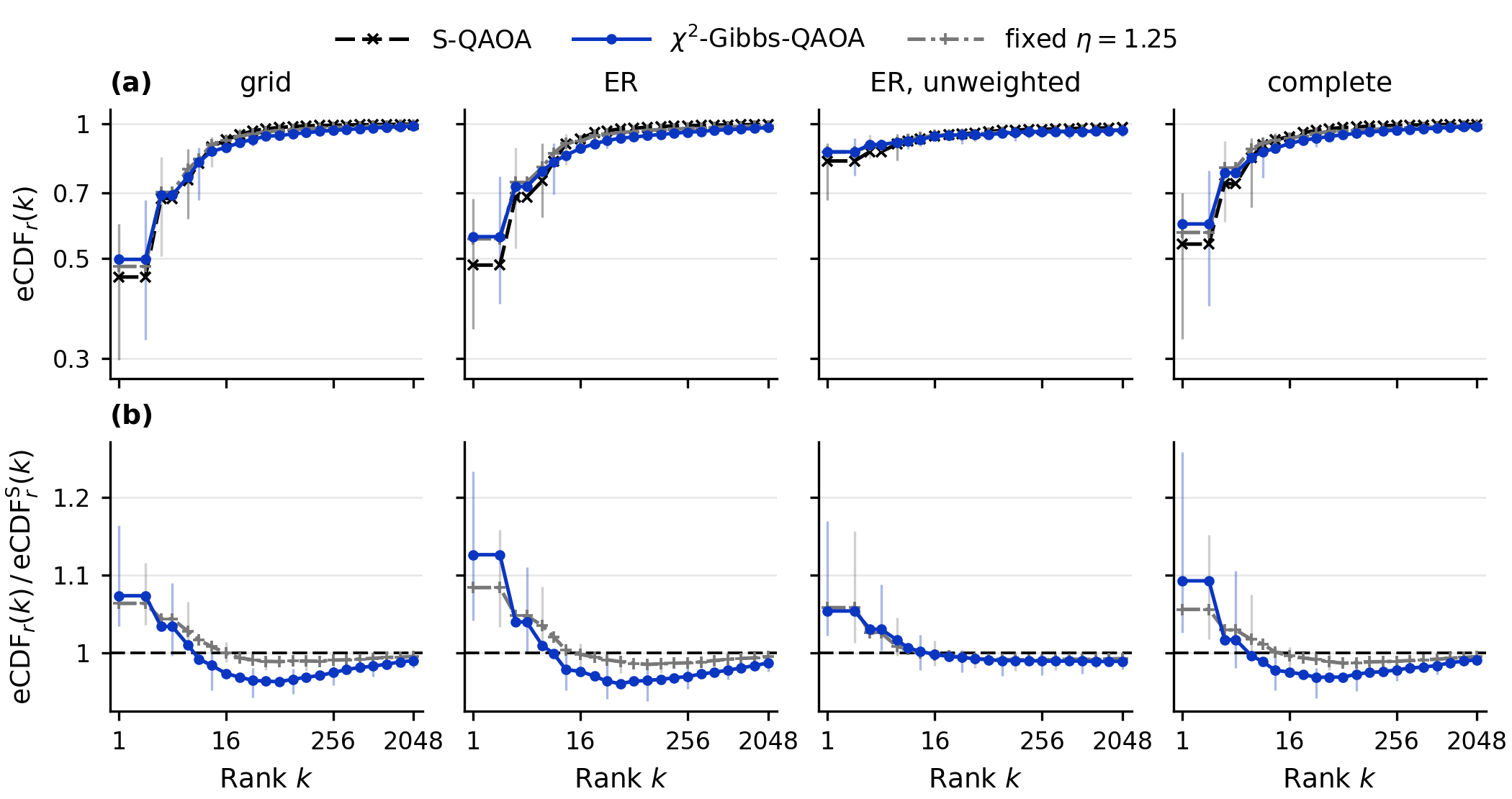}
\caption{Same as Fig.~\ref{fig:s16-ecdf} at $p=32$.}
\label{fig:s16-p32}
\end{figure*}

Figs.~\ref{fig:s16-p16} and~\ref{fig:s16-p32} show the 16-qubit families of Fig.~\ref{fig:s16-ecdf} at $p=16$ and 32. The three weighted families keep similar profiles at these depths, and the unweighted ER keeps the highest ground-state probability.

\section{Details of the iterative interpolation study}
\label{app:IIA}

\subsection{Algorithm and parameters}
\label{app:dl}

We use the iterative interpolation algorithm of Ref.~\cite{apte2026iterative} (Algorithm~1 therein) with the Chebyshev basis. In the notation of that reference, the starting depth is $p_0=5$, the maximum depth $p_{\max}=300$, the initial number of coefficients $\mathcal{C}=4$ per schedule, the improvement threshold $\epsilon=10^{-3}$ and the patience $\tau=1$, with no target approximation ratio. Two points differ from Ref.~\cite{apte2026iterative} however. The objective function is estimated from $K$ shots, as in Sec.~\ref{sec:constant-depth}, and the depths are not equally spaced (Appendix~\ref{app:adiabatic}).

\begin{figure*}[t!]
\centering
\includegraphics[width=\textwidth]{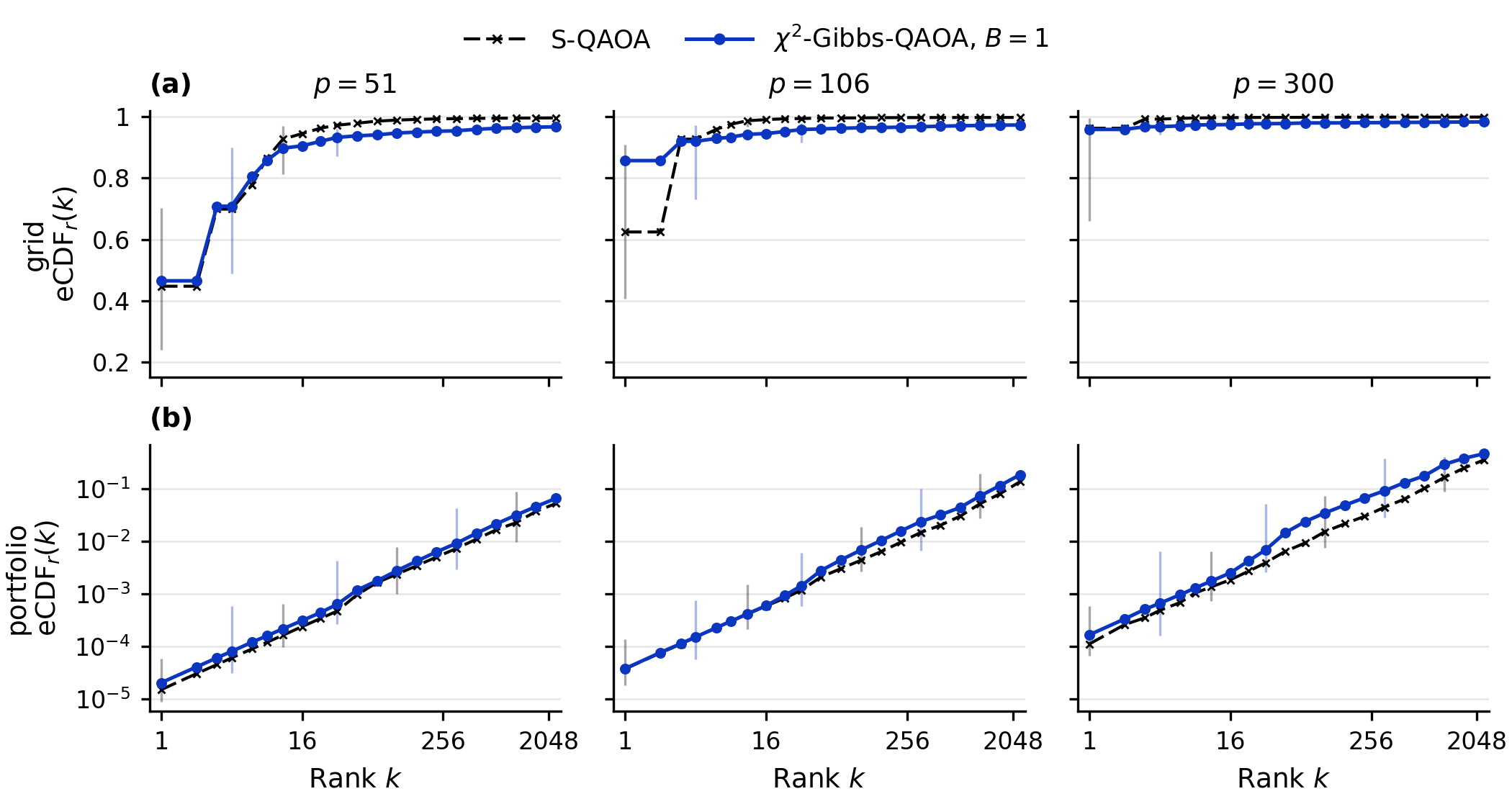}
\caption{$\mathrm{eCDF}_r(k)$ behind the ratios of Fig.~\ref{fig:dl}, $K=10^3$: (a) Max-Cut grid and (b) portfolio optimization at $p=51$, 106 and 300, for S-QAOA and $\chi^2$-Gibbs-QAOA ($B=1$). Medians over the 32 instances with interquartile bars.}
\label{fig:app-dl-ecdf}
\end{figure*}

The BOBYQA calls use $\rho_{\mathrm{beg}}=0.035$ and $\rho_{\mathrm{end}}=10^{-6}$ on the coefficients of $2\boldsymbol{\gamma}$ and $-\boldsymbol{\beta}$, a relative tolerance $10^{-8}$ and at most $10^4p$ evaluations per call, a limit never reached. The value of the objective function that decides whether a coefficient is added is estimated on $K$ fresh shots at the end of each call, with the $\eta$ of that depth for $\chi^2$-Gibbs-QAOA. Fig.~\ref{fig:app-dl-ecdf} gives the $\mathrm{eCDF}_r(k)$ behind the ratios of Fig.~\ref{fig:dl}, with $p=51$ added. As in Appendix~\ref{app:s25-results}, the ratios of these medians differ from the medians of the ratios shown in the main text. On portfolio optimization at $p=106$, for instance, the two median curves nearly coincide over the 16 lowest states, while the median of the ratios is about 1.2. On Max-Cut grid, $\chi^2$-Gibbs-QAOA leaves about ten times more weight than S-QAOA beyond the ranks shown (2.7\% against 0.25\% at $p=106$).

\subsection{Log spacing for QAOA depth increase}
\label{app:adiabatic}

Ref.~\cite{apte2026iterative} increases the depth by a constant step $\Delta p$. We use instead 52 depths in geometric progression from $p_{\min}=5$ to $p_{\max}=300$, rounded to distinct integers. Every depth from 5 to 25, and then a ratio of about 1.08 between consecutive depths (27, 29, 32, \dots, 255, 277, 300). Above $p=25$, each interpolation thus changes the depth by the same relative amount, and the number of optimizer calls grows as $\ln p_{\max}$ instead of $p_{\max}$. The cost of an evaluation being proportional to $p$, the sum of the 52 depths is 3880, against 9150 for the 60 depths of a constant step $\Delta p=5$. This choice we make is based on adiabatic theorem considerations \cite{born1928beweis,albash2018adiabatic}, where we consider  a linear ramp $\delta t(p) = t_{\mathrm{max}}/p$ and want to keep its relative evolution as constant as possible during training. This speeds up the training by a factor $\sim3$ without significantly impacting the results in our testings, which is useful for large qubit instances.

\begin{figure*}[t!]
\centering
\includegraphics[width=0.8\textwidth]{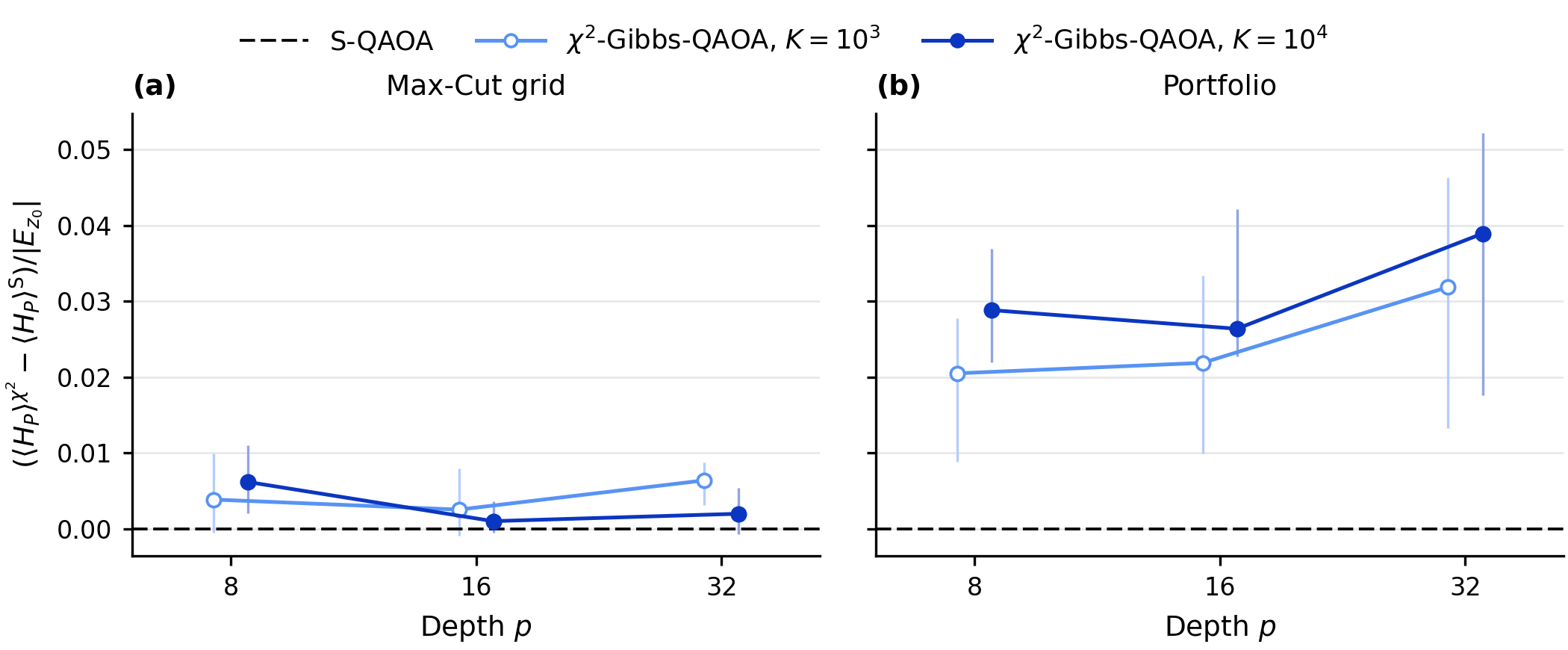}
\caption{Normalized mean energy difference between $\chi^2$-Gibbs-QAOA and S-QAOA for Max-Cut grid (left) and portfolio optimization (right) from the constant loop of Sec.~\ref{sec:constant-depth} at 25 qubits (median over 32 instances)}
\label{fig:mean-energy}
\end{figure*}
  
\end{appendix}

\end{document}